\documentclass[lettersize,journal]{IEEEtran}
\usepackage{amsmath,amsfonts}
\usepackage{algorithmic}
\usepackage{algorithm}
\usepackage{array}
\usepackage[caption=false,font=normalsize,labelfont=sf,textfont=sf]{subfig}
\usepackage{textcomp}
\usepackage{stfloats}
\usepackage{url}
\usepackage{verbatim}
\usepackage{graphicx}
\usepackage{cite}
\usepackage{scalerel}
\usepackage{tikz}
\usepackage{algorithm}
\usepackage{algorithmic}
\usepackage{graphicx}
\usepackage{subfig}
\usepackage{array}
\newcolumntype{?}{!{\vrule width 1.25pt}}
\usepackage{ctable}
\usepackage{tabulary,tabularx}
\newcolumntype{Y}{>{\centering\arraybackslash}X}
\usepackage{amsmath}
\DeclareMathOperator*{\argmax}{arg\,max}
\DeclareMathOperator*{\argmin}{arg\,min}
\usepackage{amsfonts}
\usepackage{multirow}
\usepackage{mathtools}
\usepackage[pagebackref,breaklinks,colorlinks]{hyperref}

\usepackage{adjustbox}

\usepackage[capitalize]{cleveref}
\crefname{section}{Sec.}{Secs.}
\Crefname{section}{Section}{Sections}
\Crefname{table}{Table}{Tables}
\crefname{table}{Tab.}{Tabs.}

\usetikzlibrary{svg.path}

\definecolor{orcidlogocol}{HTML}{A6CE39}
\tikzset{
  orcidlogo/.pic={
    \fill[orcidlogocol] svg{M256,128c0,70.7-57.3,128-128,128C57.3,256,0,198.7,0,128C0,57.3,57.3,0,128,0C198.7,0,256,57.3,256,128z};
    \fill[white] svg{M86.3,186.2H70.9V79.1h15.4v48.4V186.2z}
                 svg{M108.9,79.1h41.6c39.6,0,57,28.3,57,53.6c0,27.5-21.5,53.6-56.8,53.6h-41.8V79.1z M124.3,172.4h24.5c34.9,0,42.9-26.5,42.9-39.7c0-21.5-13.7-39.7-43.7-39.7h-23.7V172.4z}
                 svg{M88.7,56.8c0,5.5-4.5,10.1-10.1,10.1c-5.6,0-10.1-4.6-10.1-10.1c0-5.6,4.5-10.1,10.1-10.1C84.2,46.7,88.7,51.3,88.7,56.8z};
  }
}

\newcommand\orcidicon[1]{\href{https://orcid.org/#1}{\mbox{\scalerel*{
\begin{tikzpicture}[yscale=-1,transform shape]
\pic{orcidlogo};
\end{tikzpicture}
}{|}}}}

\usepackage{hyperref} %<--- Load after everything else
\begin{document}

\title{Robust Subgraph Learning by Monitoring\\ Early Training Representations}

\author{Sepideh Neshatfar\orcidicon{0009-0001-8896-8109}, Salimeh Yasaei Sekeh\orcidicon{0000-0002-4630-5593} 
\thanks{The authors are with the School of Computing and Information Science, University of Maine,
Orono, Maine, United States.\\
E-mails: \{sepideh.neshatfar, salimeh.yasaei\}@maine.edu}
        % <-this % stops a space
}



\maketitle
\def\ul{\underline}

\begin{abstract}
Graph neural networks (GNNs) have attracted significant attention for their outstanding performance in graph learning and node classification tasks. However, their vulnerability to adversarial attacks, particularly through susceptible nodes, poses a challenge in decision-making. The need for robust graph summarization is evident in adversarial challenges resulting from the propagation of attacks throughout the entire graph. In this paper, we address both performance and adversarial robustness in graph input by introducing the novel technique SHERD ({\ul{S}}ubgraph Learning \ul{H}ale through \ul{E}arly Training \ul{R}epresentation \ul{D}istances). SHERD leverages information from layers of a partially trained graph convolutional network (GCN) to detect susceptible nodes during adversarial attacks using standard distance metrics. The method identifies "\emph{vulnerable (bad)}" nodes and removes such nodes to form a robust subgraph while maintaining node classification performance. 
Through our experiments, we demonstrate the increased performance of SHERD in enhancing robustness by comparing the network's performance on original and subgraph inputs against various baselines alongside existing adversarial attacks. 
Our experiments across multiple datasets, including citation datasets such as Cora, Citeseer, and Pubmed, as well as microanatomical tissue structures of cell graphs in the placenta, highlight that 
SHERD not only achieves substantial improvement in robust performance but also outperforms several baselines in terms of node classification accuracy and computational complexity.
\end{abstract}

\begin{IEEEkeywords}
Subgraph Learning, Graph Neural Networks, Robustness, Node Classification, Early Training Representation
\end{IEEEkeywords}

\section{Introduction}
\label{sec:intro}
\IEEEPARstart{G}{raph} Neural Networks (GNNs) have inevitably made their way into numerous applications, such as Social Network Analysis~\cite{benslimane2023text}, Drug Discovery~\cite{wu2023deepcancermap}, and Computer Vision~\cite{munir2023mobilevig},~\cite{vanea2022new},~\cite{jung2023devils}. One of the key strengths of GNNs lies in their capacity to aggregate information around each node, contributing to superior performance compared to other machine learning methods. However, the aggregation process comes with the challenge of vulnerability propagation, leading to a reduction in performance. %significant computational complexity and challenges in terms of running time.
Graph Summarization (aka subgraph learning) has emerged as a promising technique for reducing the computational complexity of GNNs while retaining the essential structural and semantic information of the graph. There have been some works trying to summarize input graphs while maintaining performance in downstream graph classification tasks (\cite{10164013},~\cite{neshatfar2023promise}).
%and a few in node classification. 
GIB~\cite{wu2020graph} is a widely recognized graph representation learning model that integrates features and graph structure. Despite not emphasizing robustness, the authors of~\cite{wu2020graph} argue for inherent robustness against attacks, but they do not explicitly address enhancing input-level robustness in GNNs. %GIB~\cite{wu2020graph} is a widely recognized work in graph representation learning, based on an information-theoretic model that encompasses features and graph structure in graph-structured data. Although this approach does not emphasize the robustness of the representation, the authors contend that the outcomes are inherently more robust against poisoning and evasion attacks when compared to other representation learning methods. However, the authors do not explicitly address enhancing robustness at the input level of GNNs (pre-processing).\\
%*********************
% Several methods focus on subgraph learning while concurrently preserving performance at this level. For instance, [\textcolor{red}{Here}]NeuralSparse~\cite{zheng2020robust} defines the input graph by sampling edges from a learned distribution to enhance performance. Despite these sparsification techniques upholding performance by learning task-relevant edges, they fail to maintain performance in the presence of node malfunctions.% and do not improve efficiency in resource consumption.\\
% This vulnerability of GNNs to adversarial attacks becomes evident when adversarial downstream information is introduced into GNNs. This Scenario facilitates the propagation of attacks through the graph, consequently undermining the learning process.
%*********************
In the majority of graph summarization works, vulnerable nodes continue to propagate attacks throughout the entire graph, causing significant harm to the adversarial performance of downstream node classification tasks. This becomes particularly evident when considering modification attacks such as DICE (Delete Internally Connect Externally)~\cite{waniek2018hiding} manipulating edges based on class labels; NEA (Network Embedding Attack)~\cite{bojchevski2019adversarial} generating adversarial perturbations that poison the graph structure; %STACK (Strict Black-box Attack)~\cite{Xu2020QueryfreeBA} a constrained optimization problem associated with the graph filter; %Removed from table
FLIP ~\cite{Bojchevski2018AdversarialAO} flipping edges according to the nodes' degrees; PGD~\cite{Madry2017TowardsDL} training a surrogate model to conduct transfer attacks. 
Additionally, unsupervised graph-improving methods (~\cite{entezari2020all}, ~\cite{wu2019adversarial},~\cite{zheng2020robust}) solely consider structural enhancement and are based on fundamental input information. To address vulnerabilities and overcome limitations associated with incomplete graph information %these two issues 
, it is imperative to identify and eliminate susceptible nodes. This process is essential for constructing a resilient subgraph that results in high-performing node classification. Accurate identification of these nodes can significantly enhance GNN performance under various adversarial attacks. To the best of our knowledge, no current method tackles the challenge of identifying a robust subgraph at the input level while maintaining the original performance of the node classification task. It is noteworthy that other approaches like~\cite{wang2023brave,chowdhury2023sparse} prune both the input graph and GNN concurrently. This presents a significant drawback in terms of the generalizability of GNNs. Moreover,~\cite{zhang2020gnnguard,chang2021not,geisler2021robustness} prioritize robustness through GNN modification in their approach, a facet beyond the scope of our current study.\\
In this study, we introduce a novel approach, \emph{SHERD (\textbf{S}ubgraph Learning that is \textbf{H}ale through \textbf{E}arly Training \textbf{R}epresentation \textbf{D}istances}), which leverages partially trained representations of both vanilla and adversarially attacked inputs. The representations are computed using a graph convolutional network (GCN) with a single hidden layer~\cite{kipf2016semi} for node classification tasks. This method identifies and masks out the most susceptible nodes while preserving performance. We assess the susceptibility of node clusters using standard distance metrics between partially trained representations of the vanilla input and the input when that specific cluster is under attack. We employ balanced K-Means~\cite{macqueen1967some} for clustering the nodes. Additionally, to mitigate performance loss after removing a cluster, we compare the representation of the vanilla input with the representation when that cluster is absent in the graph. The decision to retain or discard a cluster depends on the trade-off between the distances between its adversarial and absent representation from the original representation in terms of robustness and performance, respectively. This approach efficiently consumes time and memory resources by 
%1) utilizing early-layer representations of the partially trained network 
1) utilizing early trained representations of the network
and 2) training a GCN with the resulting robust subgraph, thereby conserving memory resources in both training and testing compared to the original input. Our experiments demonstrate its outperformance across various attack types and datasets. Section~\ref{Experimental_study_section} highlights the superiority of our method, particularly in the face of powerful attacks or in the absence of any attack. This underscores SHERD's effectiveness in identifying and eliminating nodes that compromise both adversarial and intact performance.\\
{\bf Our contributions} in this paper are outlined as follows:\\
(1) To the best of our knowledge, SHERD is the first approach that simultaneously addresses the crucial learning challenges of \emph{robustness} and \emph{performance} %, and \emph{efficiency} 
 for graph node classification; (2) to robustify graph node classification learning, we utilize representations of a partially trained GCN to assess the impact of retaining each set of nodes being under attack; (3) the resulting subgraph not only maintains the GCN performance with vanilla input but also enhances baseline performance with the original full graph input; and (4) our SHERD method considers efficiency when finding the subgraph.%from both intra-method and extra-method perspectives. Within the method, 
This is accomplished by partially training the network for a limited number of epochs in Phase I and calculating distances for the subsets using the hidden layer's representations. %In the evaluation process, the degree of subgraph in this approach is adjustable, with more promising results seen at some compression ratios. 
SHERD's subgraph effectively conserves memory resources in loss computation.
% \begin{itemize}
%     \item To the best of our knowledge, SHERD is the first approach that simultaneously addresses three crucial machine learning challenges of \emph{robustness}, \emph{performance}, and \emph{efficiency} for GNN input.
%     \item To validate the robustness of the subgraph derived from the GNN input, this method employs the attacked representation of a partially trained network to assess the impact of retaining each set of nodes being under attack.
%     \item The resulting subgraph not only maintains but also enhances baseline performance, achieved by comparing representations with and without the set of nodes under investigation.
%     \item The proposed method prioritizes efficiency from both intra-method and extra-method perspectives. Within the method, this is accomplished by 1) partially training the network for a limited number of epochs in Phase I and 2) calculating distances for the subsets using early network representations. After finding the final subgraph, in the evaluation process, the degree of compression in this approach is adjustable, with more promising results seen at some compression ratios. This compressed subgraph effectively conserves resources.
% \end{itemize}
\subsection{Related Work} 
% \textbf{Graph Summarization.}
% \textbf{Adversarial Defense.}
The excessive vulnerability of GNNs when under attack has been a concern addressed in various defense methods (~\cite{mujkanovic2022defenses, zheng2021graph}). These defense strategies encompass approaches: RobustGCN~\cite{zhu2019robust}, employs variance-based attention weights for aggregating node neighborhoods; GRAND~\cite{feng2020graph}, relies on stochastic graph data augmentations; ProGNN~\cite{jin2020graph}, concurrently learns graph structure and GNN parameters; GAME~\cite{zhang2023chasing}, utilizes reinforcement learning to preprocess input; GNNGaurd~\cite{zhang2020gnnguard}, exploits neighbor relevance and memory usage; NOSMOG~\cite{tian2023learning}, is built upon position features, representational similarity distillation, and adversarial feature augmentation; G-RNA~\cite{xie2023adversarially}, automatically determes optimal defensive strategies for perturbed graphs; and  Soft-Median-GDC~\cite{geisler2021robustness}, grounded in robust message-passing aggregation and draws from advancements in differentiable sorting. These methods improve GNN robustness, yet they often involve time-consuming procedures and nontransferable strategies when employing a different GNN and attacks. Hence, unsupervised graph-improving methods have emerged, focusing on preprocessing graph datasets to mitigate the impact of attacks on input.\\
Recent advancements have led to a few methods for GNN robustness by preprocessing data structures. For instance, SVD-GCN~\cite{entezari2020all} takes advantage of the inherent characteristics of input data. It proposes a technique based on the intuition that NETTACK~\cite{zugner2018adversarial}, a well-established attack method, is likely to leave perturbation traces on smaller singular values in the SVD of the adjacency matrix. However, the limitation of this approach lies in its reliance on intuition validated experimentally for a specific type of attack. Moreover, Jaccard-GCN~\cite{wu2019adversarial} assumes that all attack methods involve connecting nodes with different edges to deceive the network. This method uses Jaccard distance to quantify node similarity and make decisions about retaining or removing edges. The limitations of these two graph-improving methods lie in their exclusive focus on refining the graph's structure. 
Furthermore, 
% a recent study achieves significant compression of massive serving graph datasets by substituting training features and the graph with a limited number of virtual nodes~\cite{si2022serving}. However, 
these approaches have a drawback as they overlook the robustness of the resultant graph.
%_____________________________________________________________
\section{Problem Formulation}
\def\bX{\mathbf{X}}
Given the graph $G=(V,E)$, where $V$ is the set of vertices and $E\subseteq\big\{\{u,v\}:\hspace{1mm} u,v\in V(G), u\neq v\big\}$ is the set of edges with adjacency matrix $A$. % Denote the neighborhood of vertex  $v\in V(G)$ by $N(v)$.
Consider a triple $G(V, E,\bX)$, where $\bX\in \mathbb{R}^{|V| \times d}$ is the node attribute matrix. A subgraph of the graph $G$,  $G_{sub}=(V_{sub},E_{sub},\mathbf{X_{sub}})$ is a graph with vertex set $V_{sub}\subseteq V$, the edge set $E_{sub}\subseteq E$, and feature matrix $\bX_{sub}\in \mathbb{R}^{k \times d}$, where $k<|V|$.
%\textcolor{red}{[Notation moved to SM]}
% \paragraph{Notations}
% We summarize the notations below:
% \begin{itemize}
% \item keep this paper:https://ieeexplore.ieee.org/stamp/stamp.jsp?tp=\&arnumber=9835689
% \item CKA paper: Similarity of Neural Network Representations Revisited
% \item $\sigma(.)$: activation function (usually ReLU)
% \item $\tilde{A}$: preprocessed adjacency matrix $A$
% \item $W^{(l)}$: trainable weight matrix of layer $l$
% \item $\theta=\{W^{(1)}, W^{(2)}\}$: set of all trained parameters
% \item $H^{(l+1)}$: Hidden layer $l+1$ (representation after layer $l$):
% \begin{equation}
%     H^{(l+1)}= \sigma(\tilde{A}\;H^{(l)}W^{(l)}).
% \end{equation}
% \item $\Delta$: Set of all possible subgraphs of the graph $G$
% \item $U^{adv}$: A possible subgraphs of the graph $G$ under attack
% \item $d_R(.)$: dobustness distance function
% \item $d_P(.)$: performance distance function
% \item $\Delta$
% \item CompRatio: ratio of input that is eliminated
% \item Del: DeltaCon, Ham: Hamming, Jac: Jaccard, Lap: LaplacianuSpectral distance metrics
% \end{itemize}
Here, we consider the node classification problem where class $y\in\mathcal{Y}$ is assigned to node $v$, $\mathcal{Y}$ is class space, and $Y$ denotes class (label) variable. 
%For the node classification problem, following~\cite{kipf2016semi}, 
Consider a single hidden layer GCN, and denote $logsof$ to be $log\textnormal{-}softmax$ function, $W^{(l)}$ the weight matrix of layer $l$,  $\theta=\{W^{(1)}, W^{(2)}\}$ the set of all trained parameters, 
%\textcolor{red}{[I have replaced all $\tilde{A}$s with A]
% $A$ adjacency matrix, 
% $\tilde{A}$ the preprocessed adjacency matrix $A$, 
 and $\sigma(.)$ activation function (usually ReLU). 
%The output of the hidden layer is\textcolor{red}{
Given that $Z$ represents the output of the GCN, the loss function is as given as $\ell(Z,Y)=\text{NLL-loss}(Z,Y)$
% \begin{equation}
%    {Z}= f_{\theta}(A,X)=logsof(\tilde{A}\;\sigma(\tilde{A}\mathbf{X}W^{(1)})\;W^{(2)}). 
% \end{equation}
% \begin{equation}
%    {Z}=f_{\theta}(A,X)=logsof(A\;\sigma(A\mathbf{X}W^{(1)})\;W^{(2)}). 
% \end{equation}
% \begin{equation}
%    \ell(Z,Y)=\text{NLL-loss}(Z,Y), 
% \end{equation}
where NLL-loss is the negative log-likelihood loss and $Y$ is the ground truth node label. 

% \begin{equation}
%    {Z}= f_{\theta}(A,X)=log\textnormal{-}softmax(\tilde{A}\;\sigma(\tilde{A}\mathbf{X}W^{(1)})\;W^{(2)})
% \end{equation}
In this research, {\it our objective is to find the most class-informative subgraph of the input data using the early training representations of GCN while preserving network performance and concurrently improving the robustness of the network against evasions}. 
%(We got the same results when exploiting later representations or summation of both representations.)
To this end, we solve an optimization problem with two major objectives: First, finding the optimal subgraph $G_{sub}$ that minimizes the difference between loss given attacked subgraph input ($U^{adv}$) and non-attacked original input $G$. Second, minimizing the distance between the loss given the vanilla original and subgraph input. % Denote $H^{(l)}$ the hidden layer $l$. %(representation after layer $l-1$).
%The optimization objective to find the robust and informative subgraph is as follows.
Therefore our optimization objective is as follows:

\begin{align}
 G_{sub} = \argmin _{U \in \Delta} \big ( \mid \ell(Z ,Y)-\ell(Z\mid _{U^{adv}} ,Y\mid _{U})\mid \nonumber
     \\  +  \mid \ell(Z ,Y)-\ell(Z\mid _{U} ,Y\mid _{U})\mid \big ),%-\mid d( G,U)\mid,
     \label{eqn:GsubOptim}   
\end{align}

% \textcolor{red}{
% \begin{align}
%  G_{sub} = \argmin _{U \in \Delta} \big ( \mid {Z}\mid _{H^{(0)}=G}-{Z}\mid _{H^{(0)}=U^{adv}}\mid \nonumber
%      \\  + \mid {Z}\mid _{H^{(0)}=G}-{Z}\mid _{H^{(0)}=U}\mid -\mid d( G,U)\mid,
%      \label{eqn:GsubOptim}   
% \end{align}
% }
% \begin{align}
%      G_{sub}
%      = \argmin _{U \in \Delta} \big (& \mid {Z}\mid _{H^{(0)}=G}-{Z}\mid _{H^{(0)}=U^{adv}}\mid \nonumber
%      \\  +&\mid {Z}\mid _{H^{(0)}=G}-{Z}\mid _{H^{(0)}=U}\mid \nonumber
%      \\  -&\mid d( G,U)\mid,
%      \label{eqn:GsubOptim}
% \end{align}
where $\Delta$ is the set of all possible subgraphs of $G$, and $d(.)$ is a standard distance metric. Note that here $\ell(Z\mid _{U} ,Y\mid _{U})$ is the loss for the nodes in subgraph $U$, and $\ell(Z\mid _{U^{adv}} ,Y\mid _{U})$ is the loss for the nodes in subgraph $U$ when they are under an adversarial attack, $U^{adv}$. This optimization helps GCN to update the weights of the network based on the most robust informative nodes of the graph. 
\section{Proposed Method: SHERD}
In this section, inspired by the recent non-graph paper~\cite{ganesh2023q}, we present a novel framework named SHERD that leverages early training representations of GCN to 
detect susceptible non-informative input nodes and eliminate them to form a robust informative subgraph. The susceptible nodes are those that fail to sustain their adversarial performance. Our methodology comprises two major phases, with 
%Phase II further divided into 
three subphases, clustering, robustness evaluation, and performance assessment, which are outlined as follows.
\subsection{Phase I} 
\label{phaseI}
In the first phase of SHERD, a standard GCN is trained on the original input for a specific number of epochs, say $\tau$. Thus, the network parameters are trained only for $\tau$ epochs with $\tau << \tau*$, where $\tau*$ is the optimum number of training epochs.
%for the GCN to train the input. 
The network parameters is the set of parameters of each layer only trained for $\tau$ epochs, $\theta_{\tau}=\{W_{\tau}^{(1)}, W_{\tau}^{(2)}\}$, and the output of the partially trained GCN is:

\begin{equation}
    {Z_{\tau}}:= f_{\theta_{\tau}}(A,\mathbf{X})=logsof(A\sigma(A\mathbf{X}\; W_{\tau}^{(1)})\; W_{\tau}^{(2)}),
\end{equation}

where $W_{\tau}^{(1)}$ and $W_{\tau}^{(2)}$ are the partially trained weights of the hidden layer and output layer, respectively. Now, the training stops, and the partially-trained frozen parameters are passed to Phase II to find $G_{sub}$
%a robust informative subgraph 
considering all three aspects, structure, nodes, and GCN parameters~\cite{si2022serving}.
\subsection{Phase II}\label{sec:phase2}
In this phase, we partition the input data into clusters using the balanced K-Means clustering technique and assess their suitability for inclusion in the final subgraph based on a scoring mechanism. Nodes in clusters deemed susceptible and trivial are pruned from the graph using these scores. The decision to utilize K-Means for clustering stems from its superior ability to separate clusters compared to methods such as random clustering, as well as its efficiency when compared to hierarchical clustering. The emphasis on balanced clusters is driven by the need for fair score calculation, where the size of each cluster directly influences its score. SHERD consists of three subphases that are brought into detail in the following.\\
%Now, our goal is to find the susceptibility-triviality score for each cluster and remove the vertices with the highest scores. 

% \begin{itemize}
%     \item \textbf{Subphase 1} 
 {\bf  Subphase 1:}  In the first subphase of Phase II, we employ the balanced K-Means clustering technique, a variant of the original K-Means algorithm~\cite{macqueen1967some}, to divide the input data into clusters. To achieve balance in K-Means clustering, we initially apply the original algorithm to identify cluster centroids. Subsequently, we compute the Euclidean distances of all nodes from these centroids, aiming to allocate an equal number of nodes to each cluster while keeping them as close to their respective centroids as possible. Consequently, if we have $B$ bunches $\{G_1,G_2,...,G_B\}\subset G$ with node sets $\{V_1,V_2,...,V_B\}\subset V$, then $\mid V_i\mid=\frac{\mid V\mid }{B}$.\\
    
{\bf Subphase 2:}
%    \item \textbf{Subphase 2}
In the preceding subphase, all nodes within each individual cluster, derived from subphase 1, share similar feature attributes. This commonality provides a valuable opportunity: the identification of one node as malicious can aid in pinpointing other nodes with similar malign intent. With our balanced clusters in hand, the next step is to determine the susceptibility score for each cluster ($G_i$) and remove the nodes of the clusters with the highest scores. This process involves subjecting each cluster to an updated version of the PGD attack~\cite{Madry2017TowardsDL}, referred to as $G_i^{PGD}=(A^{PGD_i},X^{PGD_i})$, to calculate the cluster's susceptibility score. This score is determined by analyzing the complete input representation when only that specific cluster is under attack, denoted as $G^{PGD_i}=\{G_1, ..., G_i^{PGD}, ..., G_K\}$. This initial representation of the attacked input is assessed after the hidden layer, utilizing partially trained parameters carried over from Phase I. Additionally, it can be expanded to consider representations from multiple layers, incorporating additional layer-specific parameters. To ensure a comprehensive evaluation, we modify the original PGD attack to target all nodes within the cluster, allowing each node's participation in score calculation and decision-making. The resulting adversarial representation, partially trained over $\tau$ epochs for the $i$-th cluster, is as follows    
\begin{equation}
    H^{(1)}_{\tau}\mid _{(H_{\tau}^{(0)}=X^{PGD_i})}^{A^{PGD_i}}= \sigma(A^{PGD_i}X^{PGD_i}W^{(1)}_{\tau}),
\end{equation}

% \begin{equation}
%     H^{(1)}_{\tau}\big\mid _{(H_{\tau}^{(0)}=X^{PGD_i})}^{A^{PGD_i}}= \sigma(A^{PGD_i}X^{PGD_i}W^{(1)}_{\tau}),
% \end{equation}

    %where $X^{PGD_i}$ and $A^{PGD_i}$ are the set of features and adjacency matrices of $G^{PGD_i}$ accordingly.
To be clear, we will denote $H^{(1)}_{\tau}\mid _{(H_{\tau}^{(0)}=X^{PGD_i})}^{A^{PGD_i}}$ as $H^{(1)}_{G^{PGD_i}}$. Implicitly, the initial representation of $G$ for the $i$-th cluster, in its unaltered state, can be expressed as
\begin{equation}
H^{(1)}_{\tau}\mid_{(H^{(0)}_{\tau}=X)} = \sigma(AXW^{(1)}_{\tau}),
\end{equation}
Here, $X$ and $A$ represent the feature set and adjacency matrices of the original, untouched input $G$, respectively, which we will now refer to as  $H^{(1)}_{G}$. By employing a standard distance metric, we gauge the dissimilarity between the manipulated representation and the unaltered representation. Specifically, the manipulated representation corresponds to the $i$-th cluster, enabling us to calculate the susceptibility score for this particular cluster (designated as cluster number $i$), as shown in the formula (\ref{Susceptibility score}). This susceptibility score (${S}^i_{Suscep}$), in turn, serves as the basis for computing the cluster's \emph{robustness} score as follows
%$${S}^i_{Robust} =-\mathcal{S}^i_{Suscep},$$ 
    \begin{align}
\mathcal{S}^i_{Suscep}=d_R(H^{(1)}_{G^{PGD_i}} , H^{(1)}_{G}), \label{Susceptibility score}\\
    \hbox{and}\;\;    {S}^i_{Robust} =-\mathcal{S}^i_{Suscep}, \label{RobustnessScore}
    \end{align}
    where $d_R(.)$ signifies the standard distance metric utilized in the robustness subphase. 
  %  , becomes crucial. You can find a detailed illustration of Subphase 2 in Figure~\ref{Subphase2}.

    \begin{figure}[H]
     \centering
    \subfloat{%[Robustness Measurement]{
        \includegraphics[width=1\columnwidth]{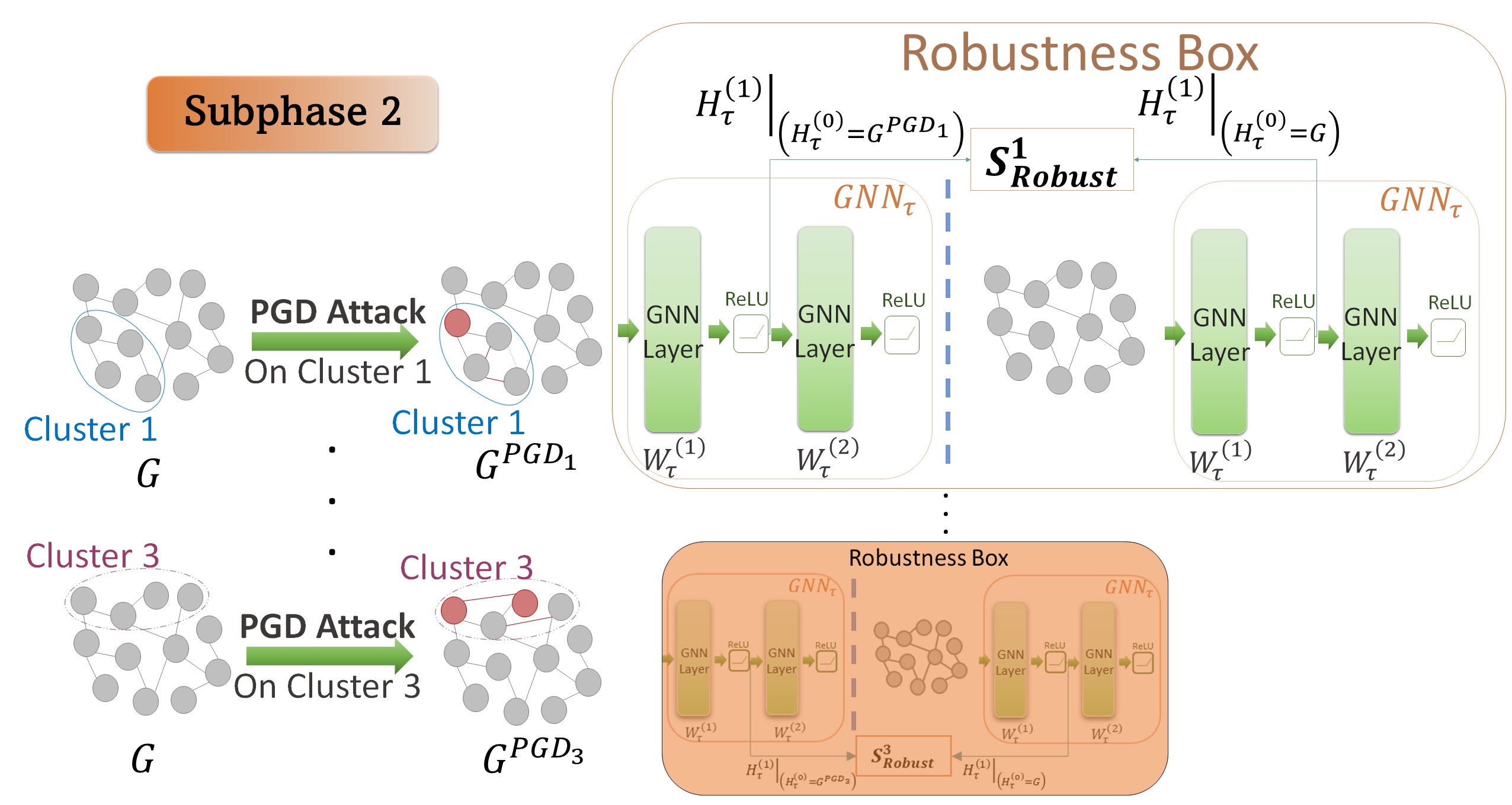}}
    % \caption{ Architecture of Subphase 2; In this figure, the process of attaining the robustness score of node clusters is visualized; The input embracing an adversarial node cluster is given to GNN for early representation retrieval. The distance between it and the similar representation for the original input results in the Robustness Score for this specific cluster. }

    \caption{Subphase2: A visualization of the process of attaining the robustness score, $\mathcal{S}_{Robust}$, of node clusters. The input embracing an adversarial node cluster is given to GCN for early representation retrieval. $\mathcal{S}_{Robust}$'s are computed based on (\ref{Susceptibility score}) and (${S}^i_{Robust} =-\mathcal{S}^i_{Suscep}$). }
    \label{Subphase2}
    % \label{Subphases}
    \end{figure}
    Since the Susceptibility Score is designed to be minimized, Equation~\ref{RobustnessScore} is introduced solely for the purpose of maximization, as we aim to solve in Equation~\ref{eqn:GsubOptim2}. For a more detailed explanation of the metric selection for both robustness and performance, see section~\ref{Metrics_Section}. A lower robustness score indicates a higher sensitivity of the cluster to attacks.  However, what if there is a cluster that, despite being sensitive to attacks, is essential for maintaining the overall graph performance? To this end, Subphase 3 (Figure~\ref{Subphase2}) is introduced to measure the performance score ($\mathcal{S}^i_{Perform}$).\\
    
\begin{figure}[h]
         \centering
    % \subfloat[Robustness Measurement]{
    %     \includegraphics[width=1\columnwidth]{Figures/Subphase2.jpg}}
    %     \label{fig:Subphase2}
    % \hfill
    \subfloat{%[Performance Measurement]{
        \includegraphics[width=1\columnwidth]{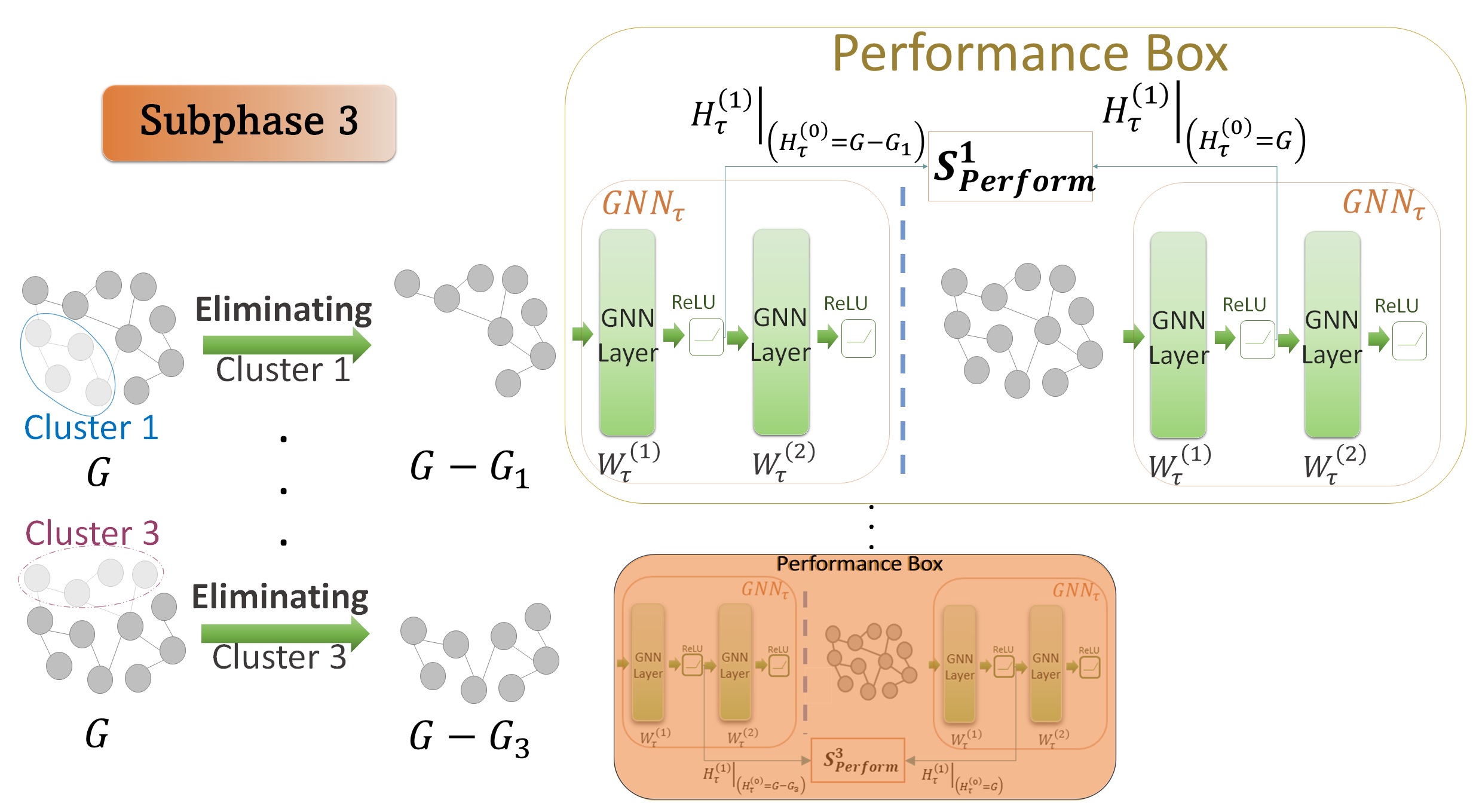}}
        
    \caption{ Subphase 3: A visualization of the process of attaining the performance score, of node clusters. The input, excluding a node cluster, is fed to the GCN for early representation retrieval. The distance between this representation and the corresponding one for the original input is $\mathcal{S}_{Perform}$ for this specific cluster.}
    
    % \label{Subphases}
    \label{Subphase3}
    \end{figure}

{\bf Subphase 3}
%    \item \textbf{Subphase 3} 
    \label{Subphase3_Item} The scores obtained in Subphase 2, applicable to all clusters, exclusively pertain to the robustness of these clusters. It is imperative to complement these scores with performance metrics that assess the significance of each cluster's presence. This involves evaluating the distance between the input representation with and w/o the specific cluster. This analysis elucidates the extent to which the representation of the entire input is influenced when each cluster is disregarded. Denote $d_P(.)$ the distance metric in performance subphase (Subphase 3, see Figure~\ref{Subphase3}), $H^{(1)}_{G-G_i}$ representation of the entire input excluding the $i$-th cluster: 
    
    \begin{equation}
        \mathcal{S}^i_{Perform}=d_P(H^{(1)}_{G-G_i} , H^{(1)}_{G}).
    \end{equation}
    % where $d_P(.)$ represents the standard distance metric employed in the performance subphase, and $H^{(1)}_{G-G_i}$ denotes the representation of the entire input excluding the $i_{th}$ cluster. A detailed illustration of this subphase is provided in Figure~\ref{Subphase3}.
    
%\end{itemize}
The ultimate decision regarding whether to retain or eliminate each group of nodes is determined by a combination of two scores, forming the robustness-performance score as defined in Equation~\ref{eqn:RobPer}.
\begin{equation}
\label{eqn:RobPer}
    \mathcal{S}^i_{Rob-Per}=\alpha \mathcal{S}^i_{Perform}+(1-\alpha) \mathcal{S}^i_{Robust}
\end{equation}

Here, $\alpha$ is a scalar in the range $[0,1]$, indicating the emphasis on the performance and robustness of each cluster. In this context, $\alpha=0.5$ assigns equal attention to both scores.
There exists a set of $B$ scores $\{\mathcal{S}^1_{Rob-Per},...,\mathcal{S}^B_{Rob-Per}\}\in\mathcal{S}_{Rob-Per}$, each representing the importance of cluster retaining. Therefore, the optimization in Equation~\ref{eqn:GsubOptim} can be reformulated as follows.
\begin{align}
     G_{sub}
     = \argmax _{U \subset \{G_1,...,G_B\}} \big ( \sum_{G_i\in U} \mathcal{S}^i_{Rob-Per} \big)%-\mid U \mid \big)
     \label{eqn:GsubOptim2}
\end{align}

\begin{figure*}
  \centering
  \includegraphics[width=1\linewidth]{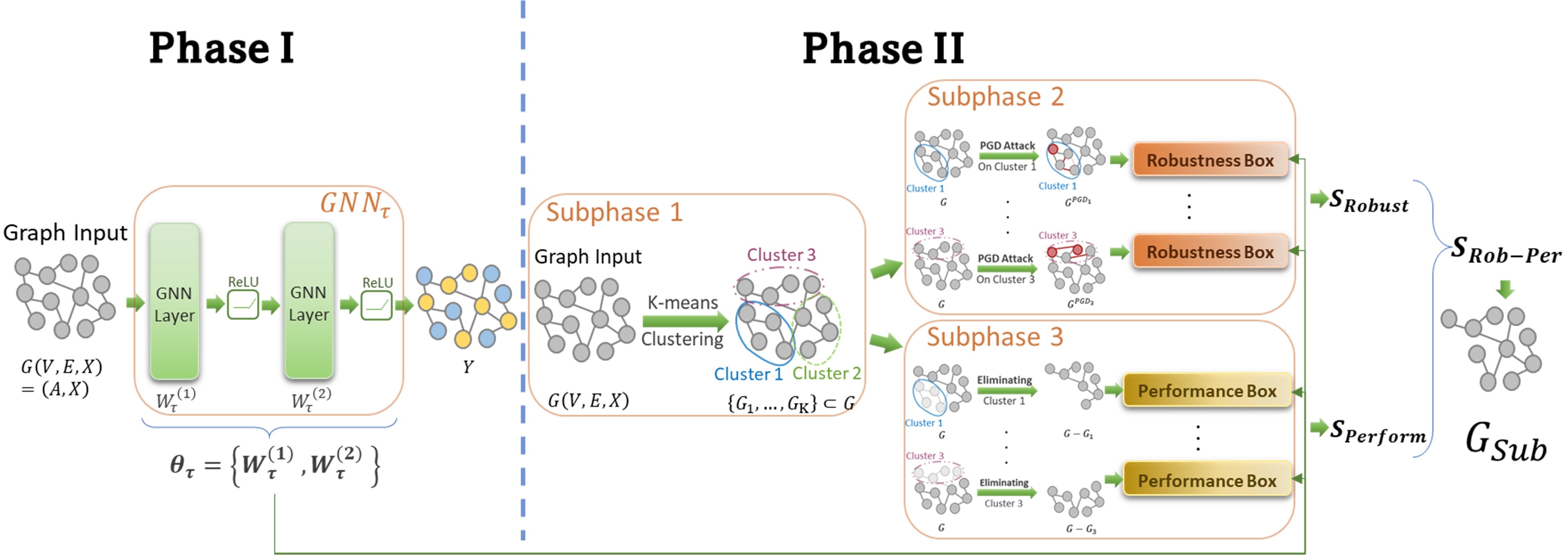}
  \caption{Overview of the entire methodology, encompassing the partial training of the GCN in Phase I and the subsequent compression process including nodes' clustering (Subphase 1), robustness analysis (Subphase 2), and performance analysis (Subphase 3) in Phase II.}
    \label{method}

\end{figure*}

This is adaptable to incorporate the user-specified compression ratio, $C$, when compressing the graph.
\begin{align}
     G_{sub}
     = &\argmax _{U \subset \{G_1,...,G_B\}} \big ( \sum_{G_i\in U} \mathcal{S}^i_{Rob-Per} \big) \nonumber \\
     &\text{s.t.} \mid U \mid =\mid G \mid \times (1-C).
     \label{eqn:GsubOptim3}
\end{align}

% The nodes are ordered based on these scores and the least $Compression Ratio \times \mid V\mid$ number of them are eliminated to form the $G_{sub}$ that satisfies formula~\ref{eqn:GsubOptim}. Here, $Compression Ratio$ is the percentage of the input graph that is intended to be dropped.

Figure~\ref{method} offers an overview of the comprehensive methodology. Additionally, refer to \ref{a1} for the detailed algorithm of this method.

\begin{algorithm}[h]
\caption{SHERD} \label{a1}
\hspace*{\algorithmicindent}\textbf{Input:\hspace{0.1mm}}{$G$,$\tau$,GCN,model,$B$,$\alpha$,$C$} 
\begin{algorithmic}[1]
\FOR {$i\in \{1,\ldots,\tau\}$}
    {\STATE\#Train  GCN on original input $G$ for $\tau$ epochs}
    {\STATE $W^{(1)} , W^{(2)}\leftarrow{\text{Train}}(GCN,G)$ }
\ENDFOR

\STATE {$\{G_1,...,G_B\} \leftarrow{\text{Balanced\_Kmeans}(G,B)}$}

\FOR {$i\in \{1,\ldots,B\}$}
    {\STATE\#Attack only being effective on $i_{th}$ cluster $G_i$}
    {\STATE $G^{PGD_i} \leftarrow{\text{PGD\_Attack}(G,mask=G_i)}$ }
    {\STATE $H^{(1)}_{G^{PGD_i}} \leftarrow{\sigma(A^{PGD_i}X^{PGD_i}W^{(1)})}$}
    {\STATE $H^{(1)}_G \leftarrow{\sigma(AXW^{(1)})}$}
    {\STATE $H^{(1)}_{G-G_i} \leftarrow{\sigma(A^{G-G_i}X^{G-G_i}W^{(1)})}$}
    {\STATE $\mathcal{S}^i_{Robust} \leftarrow{-d_R(H^{(1)}_{G^{PGD_i}},H^{(1)}_G)}$}
    {\STATE $\mathcal{S}^i_{Perform} \leftarrow{d_P(H^{(1)}_{G-G_i},H^{(1)}_G)}$}
    {\STATE $\mathcal{S}^i_{Rob-Per} \leftarrow{\alpha \mathcal{S}^i_{Perform} +(1-\alpha)\mathcal{S}^i_{Robust}}$}
\ENDFOR

{\STATE $G_{sub}$ is found using optimization \ref{eqn:GsubOptim2}}
% {\STATE \begin{align} G_{sub}= \argmax _{U \subset \{G_1,...,G_B\}} \big ( \sum_{G_i\in U} \mathcal{S}^i_{Rob-Per} \big) \nonumber \\
%      \text{s.t.} \mid U \mid =\mid G \mid \times (1-C).\nonumber \end{align}}

\textbf{Output:\hspace{1mm}}{$G_{sub}=(A_{sub},X_{sub})$} 
\end{algorithmic}
\end{algorithm}

SHERD, a novel robust-informative subgraph learning approach for inductive learning, offers several advantages:

        \textbf{1) Extensibility Across Various Dimensions}: Our approach is highly extensible across different datasets, models, and emphasis on either robustness or performance (as determined by the parameter $\alpha$ in Equation~\ref{eqn:RobPer}). It also accommodates hyperparameters such as compression ratio and training epochs, making it adaptable to diverse applications. Importantly, SHERD does not impose specific assumptions or constraints on the dataset and model, making it applicable to any real-world dataset trained using standard GCNs, unlike other preprocessing methods such as~\cite{wu2019adversarial, entezari2020all}.
        
    \textbf{2) Simplicity Without Adversarial Training and Loss Embracement}; The method stands out for its simplicity. By offline utilization of the partially trained network, it involves studying only a few parameters compared to adversarially trained methods. Despite its straightforwardness, the method proves highly effective against strong evasion attacks, as demonstrated in Table~\ref{table1}.
    
    \textbf{3) Transferability of Robustness Against Attacks}; The attack applied on each cluster during the robustness subphase of Phase II, refined PGD attack, is considered potent. So, building robustness against this method generally defends against other attacks (\cite{wiedeman2022disrupting}). Further discussion about this white-box attack is presented in Section~\ref{Attacks_Transferebility_sec}.
    
    \textbf{4) Efficiency in Time and Resource Utilization}; Partial training of the network on the original data enhances resource efficiency when identifying subgraph nodes. Rather than analyzing susceptibility and informativeness at the individual node level, the method focuses on clusters selected based on features, leveraging the inherent neighborhood information present in graph structures, where clustering naturally emerges.
    %Instead of analyzing individual nodes for susceptibility and informativeness, the method concentrates on feature-wise selected clusters, capitalizing on inherent neighborhood information in graph structures \textcolor{blue}{which is where clustering emerges}. 
    Larger cluster sizes contribute to time savings with fewer loop iterations, but the trade-off between time and memory needs consideration, as larger clusters demand more memory. Moreover, the compressed input reduces the required resources for both training and testing compared to the original input.
%_____________________________________________________________
\section{Experimental Study} \label{Experimental_study_section}
%This section utilizes figures and tables to visually demonstrate the superior performance of SHERD compared to other baseline methods.
% In this section, exploiting some figures and tables, we demonstrate how SHERD outperforms other baselines. %To the best of our knowledge, there is no other similar method that mitigates both the robustness and performance of the input of GNN when compressing it. This is why in the results we compare our method performance and robustness with the original uncompressed input and the randomly compressed one. 
%\subsection{Experimental Setup}
In our investigations, we employ the standard GCN model with a common setup featuring a single hidden layer, similar to \cite{ma2020towards} and many other works.
\begin{table*}[h!]
\caption{ Performance comparison of SHERD against baseline methods on multiple datasets under various adversarial attacks. Results report the mean test accuracy and standard deviation over multiple trials. The hyperparameters utilized for each dataset are tuned for optimal performance. SHERD consistently achieves strong improvements in accuracy and robustness compared to the original input baseline, outperforming other compression techniques.} 
\centering 
{\fontsize{8}{15}\selectfont
\setlength\tabcolsep{5pt}

\begin{tabular}{?l|l?c|c|c|c|c|c|c?c?}
\hline

\multirow{2}{*}{\textbf{Dataset}}&\multirow{2}{*}{\textbf{Method}}&\multicolumn{7}{c?} {\textbf{Attack Method}}&\multirow{2}{*}{\textbf{Vanilla}}\\
\cline{3-9}
                                && DICE&FGA&NEA&FLIP&FGSM&PGD&RND&  \\
                                % && DICE&FGA&NEA&STACK&FLIP&FGSM&PGD&RND&  \\
\toprule
\multirow{ 1}{*}{\shortstack[l]{{\bf Cora}}} &Original& .773$\pm$.007 & .825$\pm$.005 & .775$\pm$.011  & .798$\pm$.007 & .231$\pm$.095 & .437$\pm$.014 & .186$\pm$.01&.825$\pm$.005\\%& .752$\pm$.004
Nodes: 2708&   Random&.78$\pm$.02 & .82$\pm$.016 & .765$\pm$.012 &  .795$\pm$.016 & .222$\pm$.117 & .41$\pm$.032 & .231$\pm$.094&.821$\pm$.016\\%.742$\pm$.017 &
Features: 1433 &   NodeDegree& \bf{.844$\pm$.009}& .86$\pm$.005 & \bf{.847$\pm$.005 }& .835$\pm$011 &  .215$\pm$.07 &  .296$\pm$.013 & .252$\pm$.061&.86$\pm$.005\\%& \bf{.846$\pm$.006 }
 Edges: 10556 &FeaturesMean& .757$\pm$.006 & .798$\pm$.003 & .756$\pm$.005 & .77$\pm$.005 & .324$\pm$.092 & .415$\pm$.008 & .33$\pm$.069&.798$\pm$.003 \\%.717$\pm$.002 & 
 &GCNJaccard& .794$\pm$.007 & .853$\pm$.003 & .827$\pm$.003 & .782$\pm$.002 & .199$\pm$.092 & \textbf{.474$\pm$.015} & .237$\pm$.091&.853$\pm$.003 \\
 &GCNSVD& .771$\pm$.008 & .831$\pm$.005 & .814$\pm$.005 & .766$\pm$.009 & .191$\pm$.089 & .447$\pm$.017 & .196$\pm$.092&.831$\pm$.005 \\
   &\textbf{SHERD}& .827$\pm$.037& \bf{.863$\pm$.032 }& .82$\pm$.031 &  \bf{.85$\pm$.011} & \bf{ .369$\pm$.015} &  .411$\pm$.005 & \bf{.344$\pm$.009}&\bf{.864$\pm$.007} \\%.794$\pm$.038 &
\midrule

\multirow{1}{*}{\shortstack[l]{{\bf Citeseer}}} &Original& .665$\pm$.01 & .73$\pm$.012 & 0.702$\pm$.018 &  .63$\pm$.007 & .174$\pm$.001 & \textbf{.51$\pm$.007}& .186$\pm$ .012&.729$\pm$ .012\\%.65$\pm$.015 &
Nodes: 3327  &Random& .658$\pm$.02 & .722$\pm$.015 &  .67$\pm$.017 &   .598$\pm$.014 &  .195$\pm$.023 &  .473$\pm$.033& .196$\pm$.023 &.722$\pm$.015\\%.625$\pm$.019 &
Features: 3703&NodeDegree&   {\bf.731$\pm$.006} &   .763$\pm$.003 &   .738$\pm$.007 &      {\bf.697$\pm$.009} &   .221$\pm$.03 &   .409$\pm$.009&.213 $\pm$.028& .763$\pm$.003\\% {\bf.742$\pm$.004} & 
Edges: 9228& FeaturesMean&  .642$\pm$.011 & .693$\pm$.006 & .66$\pm$.009 & .572$\pm$.007 & .237$\pm$.028 & .444$\pm$.025& .228$\pm$ .012&.693$\pm$ .006\\%& .607$\pm$.005 
& GCNJaccard&  .675$\pm$.007 & .752$\pm$.002 & .731$\pm$.003 & .676$\pm$.003 & .226$\pm$.018 & .536$\pm$.009& .232$\pm$ .025&.752$\pm$ .003\\%& .607$\pm$.005 
& GCNSVD&  .671$\pm$.004 & .747$\pm$.006 & .722$\pm$.005 & .667$\pm$.007 & .219$\pm$.002 & .525$\pm$.01& .219$\pm$ .003&.747$\pm$ .006\\%& .607$\pm$.005 
& \textbf{SHERD}&   .712$\pm$.014 &  {\bf .787$\pm$.006} & {\bf  .745$\pm$.009} &        .651$\pm$.003&   {\bf.245$\pm$.029} &   .475$\pm$.024&{\bf.238 $\pm$.035}&\textbf{.787$\pm$.006} \\%.697$\pm$.012&
\midrule

\multirow{1}{*}{\shortstack[l]{{\bf Mini-Pubmed}}} &Original& .524$\pm$.018 & .523$\pm$.002 & .466$\pm$.002 &  .466$\pm$.002 & .323$\pm$.061& .474$\pm$.008& .304$\pm$.053 &.523$\pm$.002\\
Nodes: 2000  &Random& .498$\pm$.026 & .523$\pm$.008 &   .443$\pm$.013 &  .443$\pm$.013 &  .308$\pm$.06 &   .439$\pm$.018& .309$\pm$.059 &.523$\pm$.008\\  
Features: 500&NodeDegree&  .566$\pm$.022 &  .6$\pm$.010 &  .549$\pm$.002 &     .538$\pm$.004 &   .345$\pm$.009 &   .457$\pm$.014& .345$\pm$.012& .6$\pm$.012 \\ 
Edges: 463& FeaturesMean&   .479$\pm$.01 & .506$\pm$.006 & .405$\pm$.00 &  .405$\pm$.00 & \textbf{.405$\pm$.00}& .44$\pm$.02& \textbf{.405$\pm$.00} &.506$\pm$.006 \\
& GCNJaccard&   .478$\pm$.079 & .484$\pm$.058 & .392$\pm$.079 &  .392$\pm$.079 & .279$\pm$.069& .425$\pm$.06& .285$\pm$.063 &.484$\pm$.058 \\
& GCNSVD&   .484$\pm$.056 & .484$\pm$.042 & .392$\pm$.057 &  .392$\pm$.057 & .267$\pm$.058& .432$\pm$.042& .281$\pm$.060 &.485$\pm$.042 \\
& \textbf{SHERD}&  {\bf.596$\pm$.008} &  \textbf{.626$\pm$.005} &  \textbf{.552$\pm$.004} &     {\bf .563$\pm$.005} &    .383$\pm$.174 &   {\bf.53$\pm$.01}& .374$\pm$.167& \textbf{.628$\pm$.004} \\ 

\midrule

\multirow{1}{*}{\shortstack[l]{{\bf Mini-Placenta}}} &Original& .678$\pm$.024 & .804$\pm$.004 & .632$\pm$.03  & .653$\pm$.031 & .532$\pm$.042& .759$\pm$.028& .532$\pm$.042 &.804$\pm$.003\\%& \textbf{.654$\pm$.009}
Nodes: 5000  &Random& .675$\pm$.039 & .811$\pm$.015 &   .621$\pm$.037 &  .643$\pm$.038 &  .541$\pm$.034 &   .745$\pm$.029& .541$\pm$.034 &.811$\pm$.016\\%.639$\pm$.023 &  
Features: 64&NodeDegree&  .684$\pm$.031 &  .775$\pm$.006 &  .617$\pm$.01 &     .64$\pm$.003 &   .532$\pm$.04 &   .697$\pm$.036& .53$\pm$.037& .775$\pm$.004 \\%  .613$\pm$.033 & 
Edges: 2500& FeaturesMean&  .670$\pm$.025 & .78$\pm$.007 & .634$\pm$.007 & .648$\pm$.007 & .506$\pm$.031& .713$\pm$.026& .506$\pm$.031 &.782$\pm$.007 \\%.642$\pm$.015 & 
& GCNJaccard&  .668$\pm$.069 & .808$\pm$.005 & .604$\pm$.104 & .617$\pm$.099 & .555$\pm$.04& .774$\pm$.009& .564$\pm$.022 &.808$\pm$.006 \\%.642$\pm$.015 & 
& GCNSVD&  .668$\pm$.009 & .816$\pm$.003 & .626$\pm$.003 & .654$\pm$.003 & .553$\pm$.00& .761$\pm$.016& .553$\pm$.00 &.817$\pm$.003 \\%.642$\pm$.015 & 
& \textbf{SHERD}&  {\bf.747$\pm$.066} &  \textbf{.891$\pm$.001} &  \textbf{.678$\pm$.036} &    {\bf .733$\pm$.001} &   \textbf{ .686$\pm$.00} &   {\bf.831$\pm$.037}& {\bf.686$\pm$.00}& \textbf{.892$\pm$.001} \\

% \midrule

% \multirow{}{}{\shortstack[l]{{\bf Indian Pines}}} &Original& .629$\pm$.019 & .689$\pm$.024 & .69$\pm$.024  & .678$\pm$.028 & .159$\pm$.036& .388$\pm$.025& .161$\pm$.037 &.69$\pm$.025\\%& \textbf{.654$\pm$.009}
% Nodes: 5000  &Random& .589$\pm$.022 & .629$\pm$.023 &   .626$\pm$.023 &  .622$\pm$.023 &  .156$\pm$.019 &   .395$\pm$.01& .153$\pm$.021 &.63$\pm$.023\\%.639$\pm$.023 &  
% Features: 64&NodeDegree&  .684$\pm$.031 &  .775$\pm$.006 &  .617$\pm$.01 &     .64$\pm$.003 &   .532$\pm$.04 &   .697$\pm$.036& .53$\pm$.037& .775$\pm$.004 \\%  .613$\pm$.033 & 
% Edges: 2500& FeaturesMean&  .670$\pm$.025 & .78$\pm$.007 & .634$\pm$.007 & .648$\pm$.007 & .506$\pm$.031& .713$\pm$.026& .506$\pm$.031 &.782$\pm$.007 \\%.642$\pm$.015 & 
% & GCNJaccard&  .614$\pm$.027 & .686$\pm$.024 & .692$\pm$.024 & .662$\pm$.024 & .195$\pm$.041& .5$\pm$.035& .209$\pm$.037 &.688$\pm$.023 \\%.642$\pm$.015 & 
% & GCNSVD&  .548$\pm$.036 & .609$\pm$.033 & .612$\pm$.034 & .589$\pm$.034 & .231$\pm$.00& .508$\pm$.021& .232$\pm$.00 &.609$\pm$.034 \\%.642$\pm$.015 & 
% & \textbf{SHERD}&  {\bf.655$\pm$.021} &  \textbf{.71$\pm$.0.009} &  \textbf{.711$\pm$.010} &    {\bf .706$\pm$.008} &   \textbf{.05$\pm$.034} &   {\bf.436$\pm$.033}& {\bf.077$\pm$.077}& \textbf{.711$\pm$.009} \\%  .602$\pm$.166 &  
\bottomrule        
\hline
\end{tabular}
}
\label{table1}
\end{table*}

% \noindet
\textbf{Dataset}
In our analyses, we leverage standard datasets in the node classification benchmark \cite{sen2008collective, yang2016revisiting}: Cora \cite{mccallum2000automating}, Citeseer \cite{giles1998citeseer}, Pubmed \cite{sen2008collective}.%, and graph version of Indian Pines \cite{PURR1947}. These datasets consist of scientific publications with varying numbers of class labels, and Pubmed is derived from the PubMed database, focusing on diabetes. Due to limited resources, we downsize the Pubmed dataset to Mini-Pubmed, a smaller subset. %Indian Pines is a Hyperspectral image segmentation dataset that is converted to graph using \cite{liu2020cnn} with nodes being the image pixels and node features being the hyperspectral bands. 
Additionally, we use Mini-Placenta, a smaller subset of the Placenta dataset provided in a recent paper \cite{vanea2022new}. Mini-Placenta consists of cell graphs in histology whole slide images, utilizing 64-dimensional embeddings from the penultimate layer of the cell classifier model as node features. KNN \cite{10.1007/PL00009293} and Delaunay Triangulation \cite{guibas1992randomized} serve as proxies for cellular interaction, with edges constructed based on Euclidean distance. Following the setup of \cite{xu2018representation} and \cite{ma2020towards}, we randomly split each dataset into training (60\%), validation (20\%), and testing (20\%) sets.%\textcolor{red}{Add the new dataset}.\\

\textbf{Baselines} As there is currently no existing framework dedicated to robust subgraph learning at the pre-processing level while maintaining performance, our primary comparisons involve contrasting our method with standard compression techniques. The baselines encompass the following: \textbf{Original}, where the input graph remains untouched; \textbf{Random}, involving the random elimination of nodes from the input graph; \textbf{NodeDegree}, where input nodes are deleted based on their degrees; \textbf{FeaturesMean}, which disregards nodes with the lowest feature averages; and \textbf{SHERD}, representing our approach for robust and informative subgraph learning.
\subsection{Results}
This section is dedicated to substantiating the validity of claims regarding the robustness and informativeness of subgraphs obtained through SHERD. Table \ref{table1} presents the accuracy and standard deviation of a GCN model trained on the complete training dataset (Original) or the compressed dataset (utilizing various standard methods). Subsequently, the model is tested on both uncompressed and compressed test data subjected to adversarial attacks. We conducted 25 trials for the random selection of training input and 5 trials for all approaches in SHERD. 
% The reported results encompass 25 trials for the random selection and training of inputs (Random) and 5 trials for all other approaches, including Original.

As shown in Table~\ref{table1}, our approach outperforms all other baselines across the majority of adversarial attacks. For Cora, Citeseer, and Mini-Pubmed datasets, where graph information is predominantly conveyed through structural patterns, the NodeDegree baseline occasionally surpasses ours in a few number of attacks. This discrepancy arises because of early representations that aggregate information within a single node-hop. Conversely, in the vision data, Mini-Placenta, our method excels against all other baselines, considering both structural and feature information. \\

Additionally, Table~\ref{table2} demonstrates performance comparison between our method and baselines under Nettack~\cite{zugner2018adversarial}, as a targeted attack that considers a local picture of the graph and FG-attack~\cite{chang2020restricted} a non-smart attack having no access to the network information.
\begin{table*}[h!]
\caption{ Performance comparison under Nettack and GFA attacks.} 
\centering 
% \begin{adjustbox}{angle=90}
{\fontsize{8}{15}\selectfont
\setlength\tabcolsep{5pt}
\begin{adjustbox}{width=\textwidth}
\begin{tabular}{?p{0.9in} |l?c|c|c|c|c|c|c?}
\hline

\multirow{2}{*}{\textbf{Dataset      }}&\multirow{2}{*}{\textbf{Attack}}&\multicolumn{7}{c?} {\textbf{ Method}}\\
\cline{3-9}
                                && Original&Random&NodeDeg&FeatMean&GCNJaccard&GCNSVD&\textbf{SHERD}  \\
                                % && DICE&FGA&NEA&STACK&FLIP&FGSM&PGD&RND&  \\
\toprule 
\multirow{1}{*}{\shortstack[l]{{\bf Cora}}} &Nettack& .244$\pm$.029 & .251$\pm$.009 & .181$\pm$.033  & .258$\pm$.023 & .282$\pm$.026 & .237$\pm$.009 & \textbf{.31$\pm$.01}\\%& .752$\pm$.004
&   GFA&.825$\pm$.005 & .818$\pm$.009 & .86$\pm$.005 &  .798$\pm$.003 & .853$\pm$.003 & .831$\pm$.005 & \textbf{.864$\pm$.094}\\
&   Vanilla&.825$\pm$.005 & .818$\pm$.009 & .86$\pm$.005 &  .798$\pm$.003 & .853$\pm$.003 & .831$\pm$.005 & \textbf{.864$\pm$.094}\\
\midrule

\multirow{1}{*}{\shortstack[l]{{\bf Citeseer}}} &Nettack& .447$\pm$.012 & .447$\pm$.022 & 0.363$\pm$.016 &  .444$\pm$.016 & \textbf{.476$\pm$.014} & .454$\pm$.007& .475$\pm$ .018\\%.65$\pm$.015 &
  &GFA& .729$\pm$.012 & .704$\pm$.014 &  .779$\pm$.006 &   .707$\pm$.017 &  .752$\pm$.003 &  .747$\pm$.006& \textbf{.787$\pm$.006} \\
  &Vanilla& .729$\pm$.012 & .704$\pm$.014 &  .779$\pm$.006 &   .707$\pm$.017 &  .752$\pm$.003 &  .747$\pm$.006& \textbf{.787$\pm$.006} \\

\midrule

\multirow{1}{*}{\shortstack[l]{{\bf Mini-Pubmed}}} &Nettack& .44$\pm$.005 & .438$\pm$.002 & .445$\pm$.011 &  \textbf{.456$\pm$.016} & .402$\pm$.059& .426$\pm$.043& .441$\pm$.022 \\%.466$\pm$.002 &
 &GFA& .522$\pm$.002 & .514$\pm$.008 &   .6$\pm$.0 &  .506$\pm$.006 &  .484$\pm$.058 &   .485$\pm$.042& \textbf{.541$\pm$.016} \\
  &Vanilla& .522$\pm$.002 & .514$\pm$.008 &   .6$\pm$.0 &  .506$\pm$.006 &  .484$\pm$.058 &   .485$\pm$.042& \textbf{.541$\pm$.016 }\\
\midrule

\multirow{1}{*}{\shortstack[l]{{\bf Mini-Placenta}}} &Nettack& .804$\pm$.004 & .784$\pm$.009 & .775$\pm$.005  & .782$\pm$.007 & .808$\pm$.006& .817$\pm$.003& \textbf{.868$\pm$.007}\\%& \textbf{.654$\pm$.009}
  &GFA& .804$\pm$.004 & .784$\pm$.009 &   .775$\pm$.005 &  .782$\pm$.007 &  .808$\pm$.006 &   .817$\pm$.003& \textbf{.868$\pm$.007}\\
  &Vanilla& .804$\pm$.004 & .784$\pm$.009 &   .775$\pm$.005 &  .782$\pm$.007 &  .808$\pm$.006 &   .817$\pm$.003& \textbf{.868$\pm$.007}\\

\bottomrule        
\hline
\end{tabular}
    \end{adjustbox}
}
% \end{adjustbox}
\label{table2}
\end{table*}

\subsection{Metrics} \label{Metrics_Section} In this methodology, Phase II employs various metrics to assess the distance between representations and selects the most promising one. Specifically, Subphases 2 and 3 consider seven metrics, the details of which are elaborated here.\\

\textbf{Pearson Correlation:} This conventional metric assesses the linear relationship between two representations. We calculate correlation coefficients using both the standard method and an updated approach, denoted as SPears, where correlations are computed between the rows of the representations. The final distance is determined by averaging these coefficients. Additional details on this metric can be found in Section~\ref{SemiPear}.% the Supplementary Material.

\textbf{Hamming Distance:} This metric quantifies the proportion of disagreeing components in the two representations (denoted as Ham). While Hamming distance is a suitable initial choice for graph distance in various analyses, it only reveals a limited facet of graph network dissimilarities.

\textbf{Jaccard:} Employing this metric provides an approximation of similarity between sets. We calculate row-wise and element-wise similarities, determining the proportion of similar non-zero elements and entire rows, respectively. Element-wise and row-wise comparisons are illustrated through EJac and RJac, respectively.

\textbf{CKA (Canonical Kernel Alignment):} This similarity index assesses correspondence between hidden layers in neural networks trained with distinct random initialization. We utilize the Canonical Kernel Alignment (CKA) method with both linear (LCKA) and Radial Basis Function (RBF) kernels (KCKA), where the RBF is based on the exponential of the $L_2$ norm of the values difference.

\begin{figure}[h!]
     \centering
 \subfloat{
    \includegraphics[width=1\columnwidth]{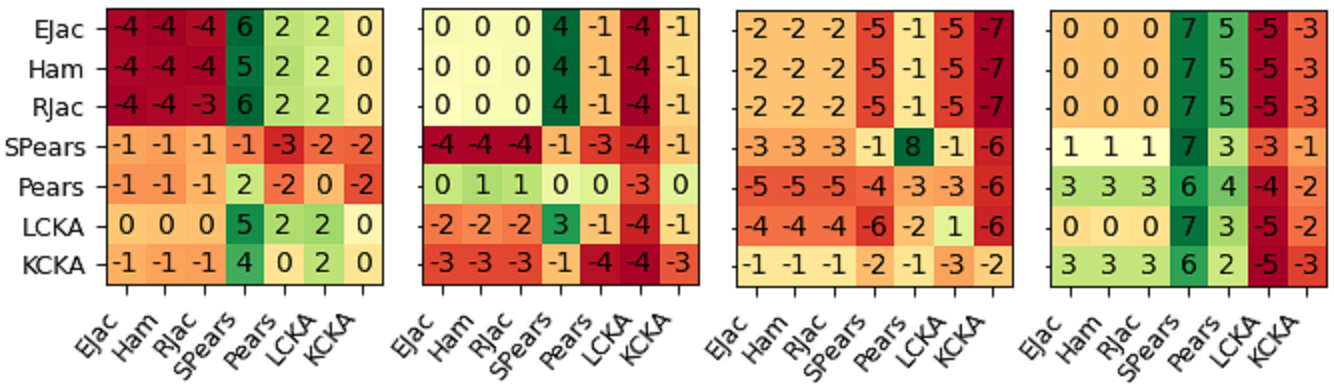}}
% \subfloat{
%     \includegraphics[width=.3\columnwidth]{Figures/DistsHeatmapCora.png}}
% % \hfill
% \subfloat{
%     \includegraphics[width=.3\columnwidth]{Figures/DistsHeatmapCiteseer.png}}
% % \hfill
% \subfloat{
%     \includegraphics[width=.3\columnwidth]{Figures/DistsHeatmapPubmed.png}}
   \caption{Distance Heatmap of Cora (first), Citeseer (second), Mini-Pubmed (third), and Mini-Placenta (fourth)}
    \label{fig:DistsHeatmap}
\label{DistsHeatmap}
\end{figure}

Figure~\ref{DistsHeatmap} visually represents the average accuracy improvement across all specified attacks for each combination of distances in both robustness ($d_R(.)$, vertical axis) and performance ($d_P(.)$, horizontal axis) subphases. The heatmap's magnitude signifies the deviation from the average accuracy of the original input for each distance combination. Hyperparameters are chosen based on optimal results, as detailed in section~\ref{Ablation_Study_Section}. Larger positive values indicate better performance of the distance compound, while negative values suggest that the baseline (original input) outperforms the specific configuration.
\subsection{Ablation Study}\label{Ablation_Study_Section}
The experiments in this paper, conducted with various sets of hyperparameters, are consolidated into a unified set showcasing the best performance for each dataset. For a comprehensive overview of all hyperparameter figures, please refer to Section~\ref{HyperParFigs}. % the Supplementary Material.
 The optimal distance combinations for the Cora, Citeseer, and Mini-Placenta datasets are either Jac or Ham for $d_R(.)$ and Semi-Pears for $d_P(.)$. In the case of the Mini-Pubmed dataset, Semi-Pears for $d_R(.)$ and Pears for $d_P(.)$ outperform other configurations, while maintaining a similar setup to other datasets still outperforms other baselines. The candidate hyperparameters include $\tau \in \{10, 30, 50, 70\}$, $B \in \{50, 100, 200, 500\}$, and $C \in \{0.1, 0.2, 0.3, 0.4, 0.5\}$, representing the proportion of data being removed. The results presented in Table \ref{table1} are obtained with the following configurations: \textbf{Cora:} $\tau=50$; B $= 200$; $C=0.4$. \textbf{Citeseer:} $\tau=70$; B $= 200$; $C=0.5$. \textbf{Mini-Pubmed:} $\tau=50$; B $= 100$; $C=0.5$. \textbf{Placenta:} $\tau=30$; B $= 100$; $C=0.5$. Additionally, $\alpha$ in Formula~\ref{eqn:RobPer} is consistently set to 0.5 across all experiments, assigning equal weights to each subphase.

\begin{figure}[h!]
     \centering
 \subfloat{
    \includegraphics[width=1\columnwidth]{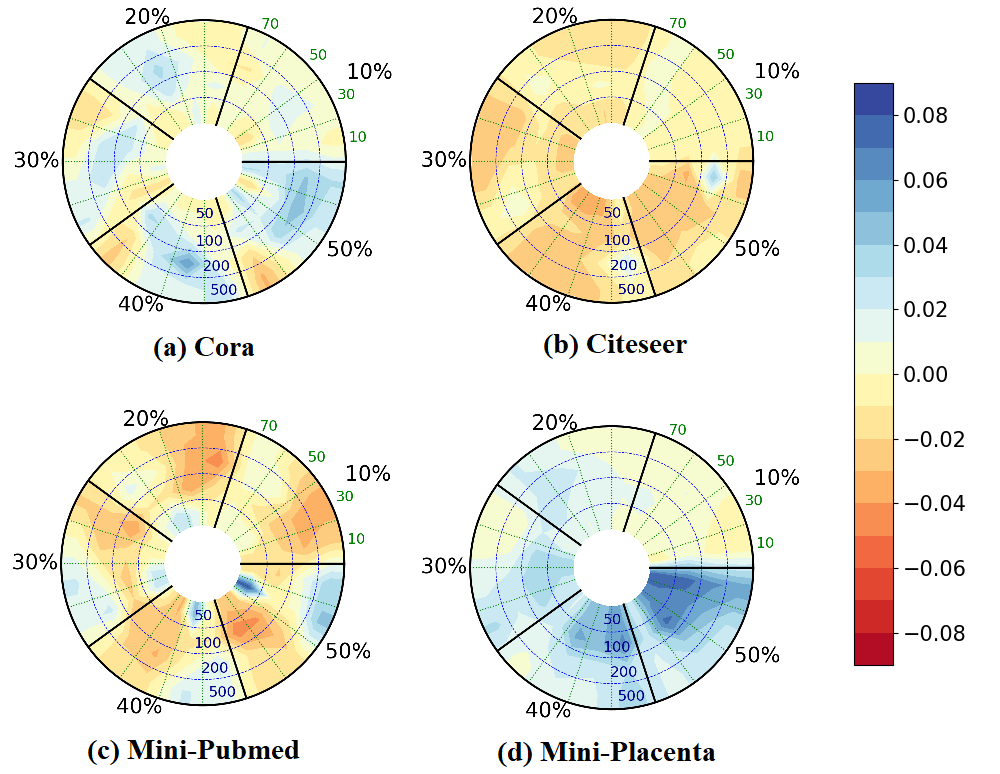}}
% \subfloat[Cora]{
%     \includegraphics[width=0.48\columnwidth]{Figures/Cora_HypParam.png}}
%     \label{fig:Cora}
% %\hfill
% \subfloat[Citeseer]{
%     \includegraphics[width=0.48\columnwidth]{Figures/Citeseer_HypParam.png}}
%     \label{fig:Citeseer}
% \subfloat[Pubmed]{
%     \includegraphics[width=0.48\columnwidth]{Figures/Pubmed_HypParam.png}}
%     \label{fig:Pubmed}
\caption{The relationship between three hyperparameters: $\tau$ (training epochs before Phase II), $B$ (number of cluster in balanced K-means' in Phase II-Subphase 1), and C (data compression percentage). The color bar indicates the extent to which each setup outperforms the original accuracy with negative values indicating the Original baseline outperforming ours.}
\label{HypParam-rel}
\end{figure}

Figure \ref{HypParam-rel} demonstrates how the choice of hyper-parameters $\tau$, $B$, and $C$ can affect average performance improvement over the Original baseline.
%____________________________________________________________
\section{Attacks Transferebility} \label{Attacks_Transferebility_sec}
In this section, we investigate the transferability of the attacks within the method and the evaluations~Section~\ref{Experimental_study_section}.
\subsection{Bunch-wise Attack in SHERD}
%{\bf Bunch-wise Attack in SHERD}
As an offline easy-to-apply approach, SHERD studies the robustness of graph nodes by means of the partially trained representation of the attacked and benign input. The choice of this attack is crucial since the decision of keeping or eliminating nodes depends on the representation of the attacked input. This means that if this method can neutralize the aforementioned attack, any other attack being transferable from it would be mitigated as well. PGD~\cite{madry2017towards}, as a potent adversary utilizing the multi-step variant, boasts the distinctive advantage of universality within the realm of first-order methods. Notably, the local maxima identified by PGD consistently exhibit similar loss values, shared across both conventionally trained networks and those trained adversarially. This makes SHERD to be robust against a spectrum of attacks reliant on first-order information. 
%This stronger multi-step version of FGSM~\cite{goodfellow2014explaining}, 
%ensures resilience against a broad array of attacks, encompassing even black-box scenarios when integrated into this framework.~\cite{madry2017towards} It is worthwhile to note that 
PGD attack is originally an attack only modifying the features of inputs. Here, similar to~\cite{zheng2021graph}, we adapt it to first randomly perturb edges of the graph. Then we optimize the features of all nodes in the cluster as target nodes by projected gradient descent.
\subsection{Evaluation Attacks}
%{\bf Evaluation Attacks }
 To elucidate the effectiveness of the proposed method, we subjected both the compressed and original graphs to several standard attacks, aiming to evaluate GCN performance under evasion attacks. %In a recent paper, very effective attacks are proposed which are called InfMax-Unif and InfMax-Norm\cite{ma2021adversarial}. This paper tries to solve an optimization problem related to the misclassification rate of the network. To do so, they assume some distributions over the network to make it intractable and solve the optimization after training the network on the train data. Their setup is tried to be realistic, so the attacker has knowledge about a limited number of nodes and no knowledge about the network. This is why outperformance on the datasets when under this attack, would demonstrate the method's superiority clearly. Additionally, there are some more effective attacks that are based on graph information learning that are brought into detail here.\\
% Degree, Betweenness, and PageRank use the node centrality scores corresponding to their names. They perturb the nodes with the highest node centrality scores. Moreover, Random Walk Column Sum (RWCS), and Greedily-Corrected RWCS (GC-RWCS) are smart black-box attack strategies proposed in \cite{NEURIPS2020_32bb90e8}. RWCS tries to select nodes with the highest importance scores that they have defined. GC-RWCS applies some heuristics to RWCS. These black-box attacks, utilizing the different graph information, can be a superb representative of other ones.\\
%In addition to these, some standard attacks from different categories of modification attacks are applied to the inputs to evaluate the method's resulting graphs.
Seven distinct modification attacks were employed for this purpose: DICE (Delete Internally Connect Externally)~\cite{waniek2018hiding} deletes edges with the same labels and adds edges with different ones; FGA (Fast Gradient Attack) \cite{chen2018fast} calculates gradients in the dense adjacency matrix with respect to the classification loss, pinpointing the most susceptible edges for perturbation. NEA (Network Embedding Attack)~\cite{bojchevski2019adversarial} is an attack originally designed for attacking Deepwalk. FLIP (Flipping attack)~\cite{bojchevski2019adversarial} follows a deterministic process: nodes are initially ranked by degrees in ascending order, and edges are then flipped from lower to higher degree nodes. FGSM (Fast Gradient Sign Method)~\cite{goodfellow2015explaining} involves linearizing the cost function around current parameter values, yielding an optimal max-norm constrained perturbation. RND (Random) \cite{zugner2018adversarial} employs a random attack strategy, exclusively altering the graph's structure. 
%____________________________________________________________
\section{Efficiency}
\begin{figure}[h!]
     \centering
\subfloat{
    \includegraphics[width=0.48\columnwidth]{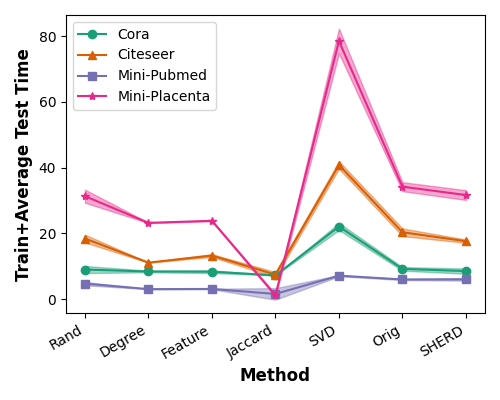}}
    \label{fig:Cora}
%\hfill
\subfloat{
    \includegraphics[width=0.48\columnwidth]{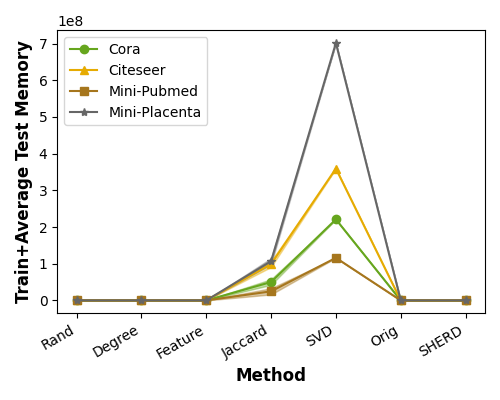}}
    \label{fig:Citeseer}
% \subfloat[Pubmed]{
%     \includegraphics[width=0.48\columnwidth]{Figures/Pubmed_HypParam.png}}
%     \label{fig:Pubmed}
\caption{Comparison of time in seconds (left) and memory usage in KiB (right) across various datasets and baselines.}
\label{Efficiency}
\end{figure}

This method exhibits superior time efficiency compared to some baselines and ranks among the best in terms of memory usage. Figure \ref{Efficiency} illustrates the memory and time usage during the training and testing of both original and subgraph inputs across various datasets, using the baselines outlined in section~\ref{Experimental_study_section}. The hyperparameters employed for our method align with those detailed in section~\ref{Ablation_Study_Section}.
%____________________________________________________________
\section{Further Distance Metric Elaboration}
\label{SemiPear}

In Phase II of this framework, diverse distance metrics are employed to calculate the Robustness (${S}^i_{Robust}$) and Performance ($\mathcal{S}^i_{Perform}$) scores for each cluster in graph $G$. These metrics comprise standard distance metrics or those derived from them. As an example, the Semi-Pearson metric is introduced in this context, which is based on the well-established Pearson correlation metric. This section provides a comprehensive explanation of the Semi-Pearson versus Pearson metric.

Suppose we have two representation matrices $H$, and $H'$ where $H, H' \in \mathbb{R}^{3\times 3}$, and each row represents a graph node embedding. Then $H=\begin{bmatrix}h_1, h_2, h_3\\\end{bmatrix}^T$,$H'=\begin{bmatrix}h'_1, h'_2, h'_3\\\end{bmatrix}^T$
% $H=\begin{bmatrix}h_1\\h_2\\h_3\\\end{bmatrix}$,$H'=\begin{bmatrix}h'_1\\h'_2\\h'_3\\\end{bmatrix}$
, and concatenating $H$ and $H'$ results in $H_C=\begin{bmatrix}h_1, h_2, h_3, h'_1, h'_2, h'_3\\\end{bmatrix}^T$.
% $H_C=\begin{bmatrix}h_1\\h_2\\h_3\\h'_1\\h'_2\\h'_3\\\end{bmatrix}$. 
The subsequent subsections compare the distance calculation for these representations using Pearson and Semi-Pearson metrics.

\subsection{Pearson Correlation Coefficients}
Given the representations $H$, and $H'$, the Pearson Correlation coefficient matrix of these two matrices is defined as:
\begin{align}
    C=\begin{bmatrix}
        1 &  R(h_1,h_2) & \cdots & R(h_1,h'_2) & R(h_1,h'_3)\\
        R(h_1,h_2) & 1 & \cdots & R(h_2,h'_2) & R(h_2,h'_3)\\
        \vdots & \vdots & \ddots & \vdots & \vdots\\
        R(h_1,h'_3)&R(h_2,h'_3)& \cdots & R(h'_2,h'_3) & 1\\
    \end{bmatrix}
\end{align}

, where $R(a,b)=\frac{Cov(a,b)}{\sqrt{Var(a).Var(b)}}$ with $Cov(.)$ representing covariance and $Var(.)$ representing variance.

Using this coefficient matrix ($C\in \mathbb{R}^{6\times 6}$), the distance is computed as the average value across all elements of this matrix.

% \subsection{Cora}

\begin{figure*}[h!]
     \centering
 \subfloat{
    \includegraphics[width=1 \linewidth]{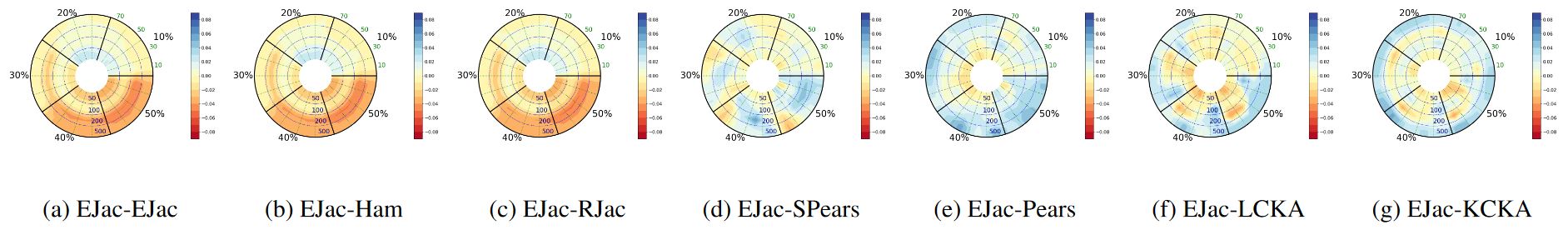}}
    
 \subfloat{
    \includegraphics[width=1 \linewidth]{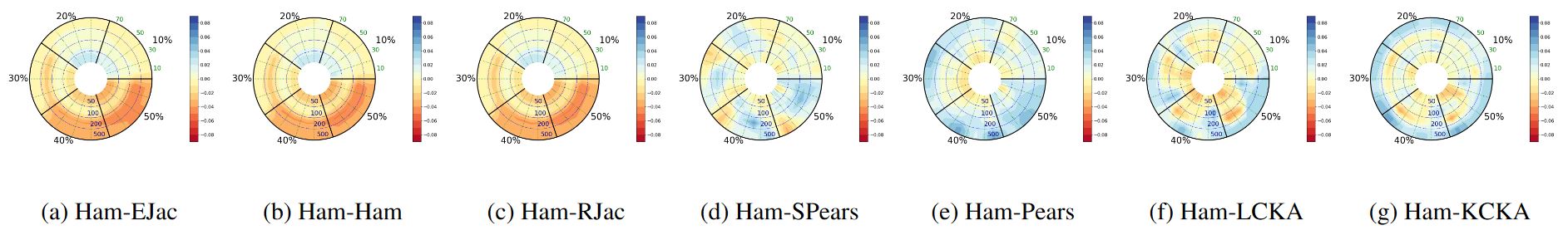}}
    
 \subfloat{
    \includegraphics[width=1 \linewidth]{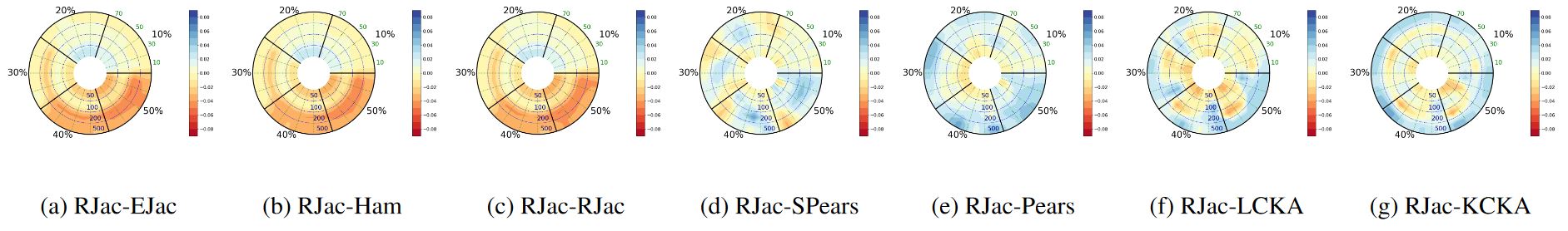}}
    
 \subfloat{
    \includegraphics[width=1 \linewidth]{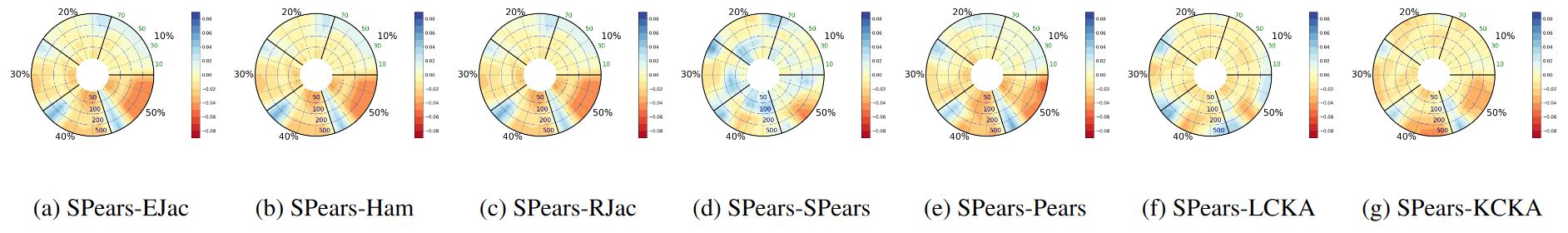}}
    
 \subfloat{
    \includegraphics[width=1 \linewidth]{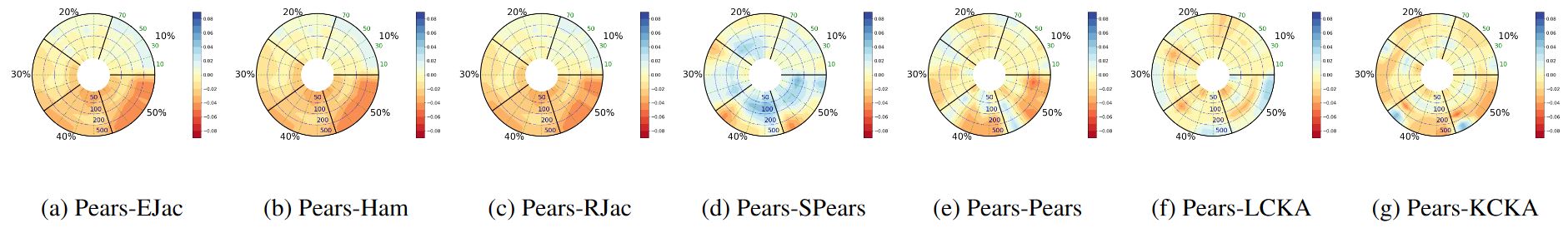}}
    
 \subfloat{
    \includegraphics[width=1 \linewidth]{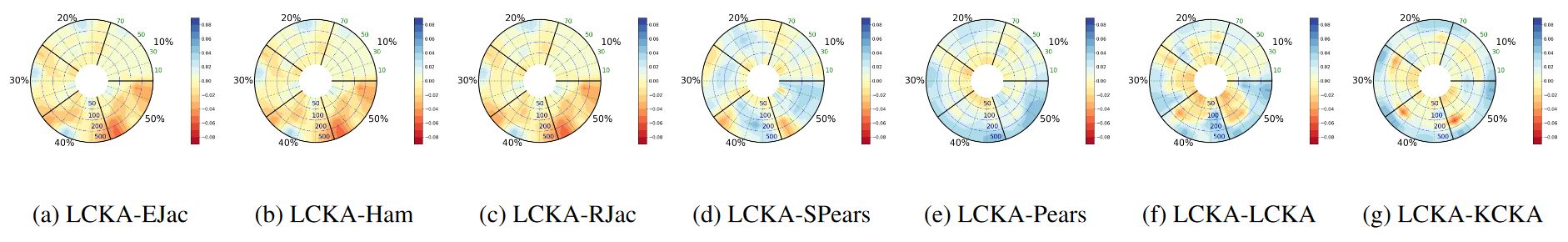}}
    
 \subfloat{
    \includegraphics[width=1 \linewidth]{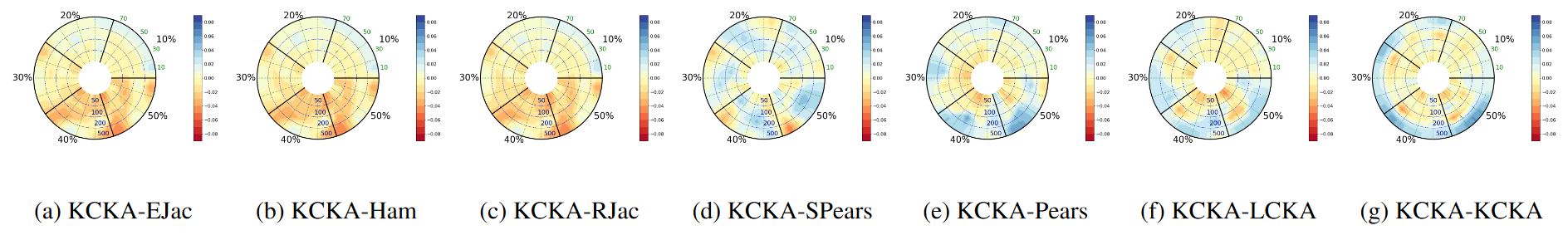}}
    \caption{Hyperparameters' relationship figures for all combinations of distance metrics for Cora dataset.}
\label{Cora}
\end{figure*}
% \newpage
% \newpage
% \pagebreak
\clearpage
\begin{figure*}[h!]
     \centering
 \subfloat{
    \includegraphics[width=1 \linewidth]{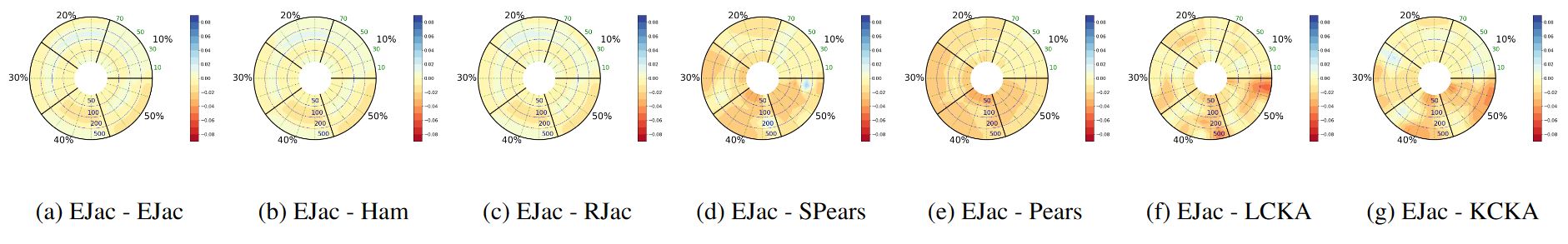}}
    
 \subfloat{
    \includegraphics[width=1 \linewidth]{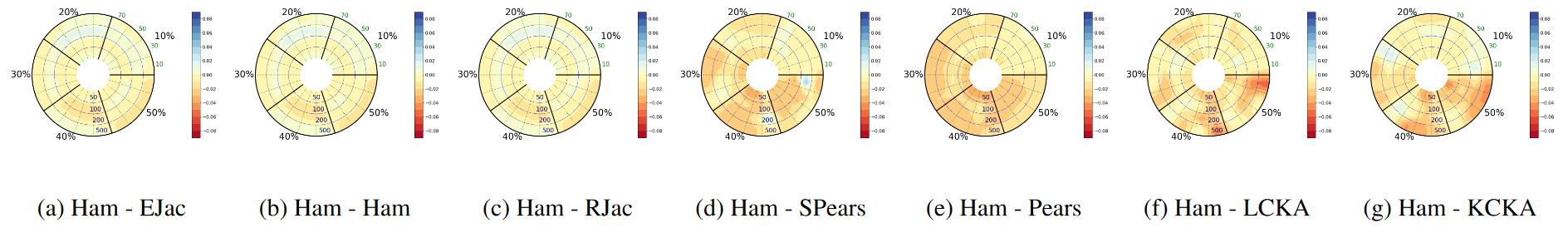}}
    
 \subfloat{
    \includegraphics[width=1 \linewidth]{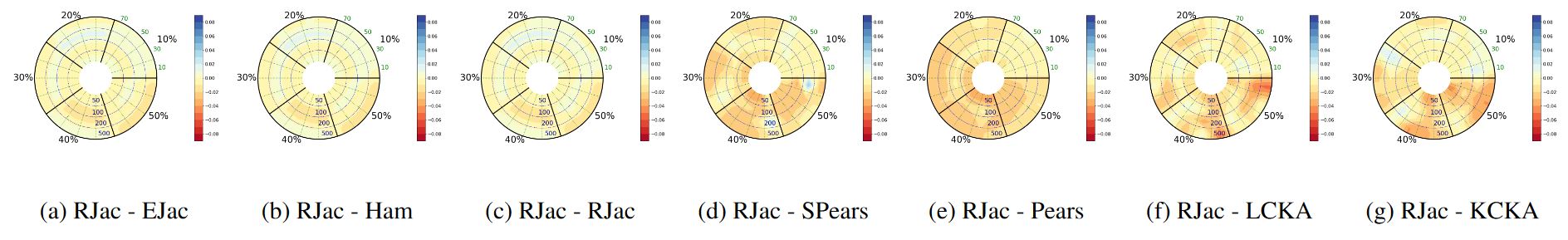}}
    
 \subfloat{
    \includegraphics[width=1 \linewidth]{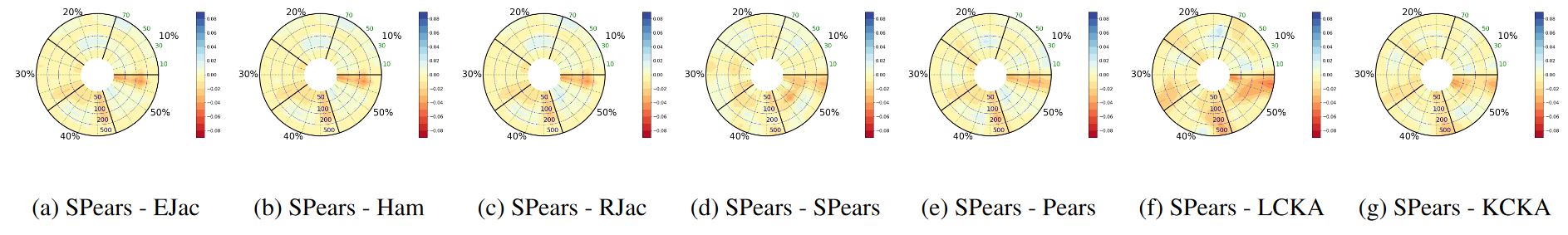}}
    
 \subfloat{
    \includegraphics[width=1 \linewidth]{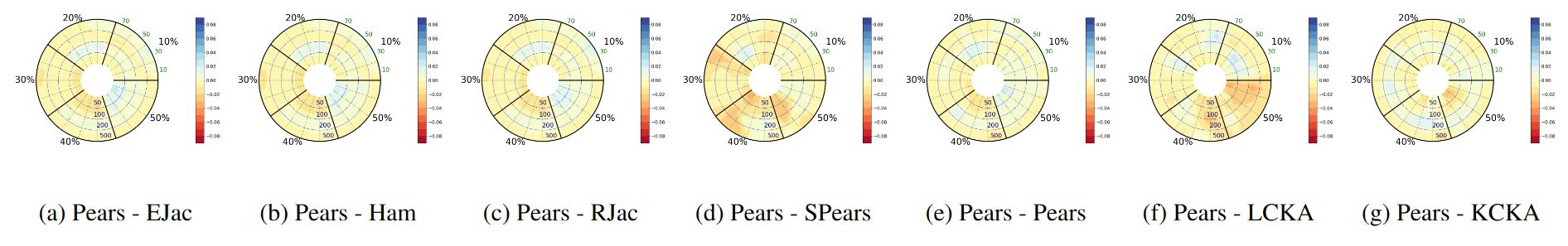}}
    
 \subfloat{
    \includegraphics[width=1 \linewidth]{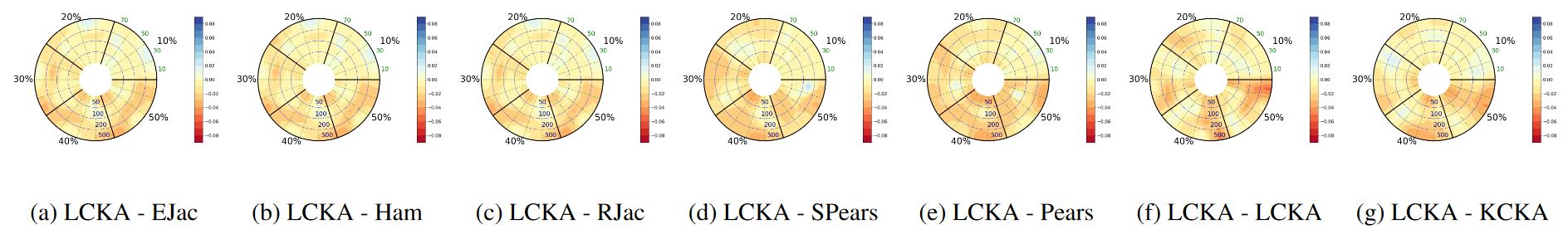}}
    
 \subfloat{
    \includegraphics[width=1 \linewidth]{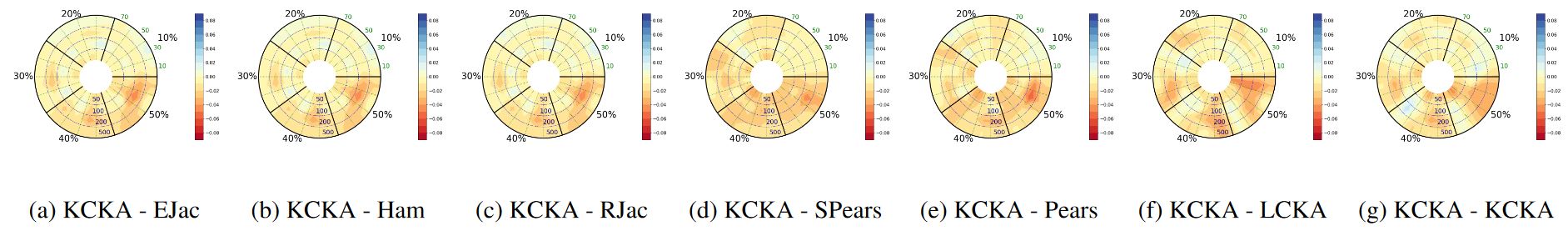}}
    \caption{Hyperparameters' relationship figures for all combinations of distance metrics for Citeseer dataset.}
\label{Citeseer}
\end{figure*}
\clearpage

\begin{figure*}[h!]
     \centering
 \subfloat{
    \includegraphics[width=1 \linewidth]{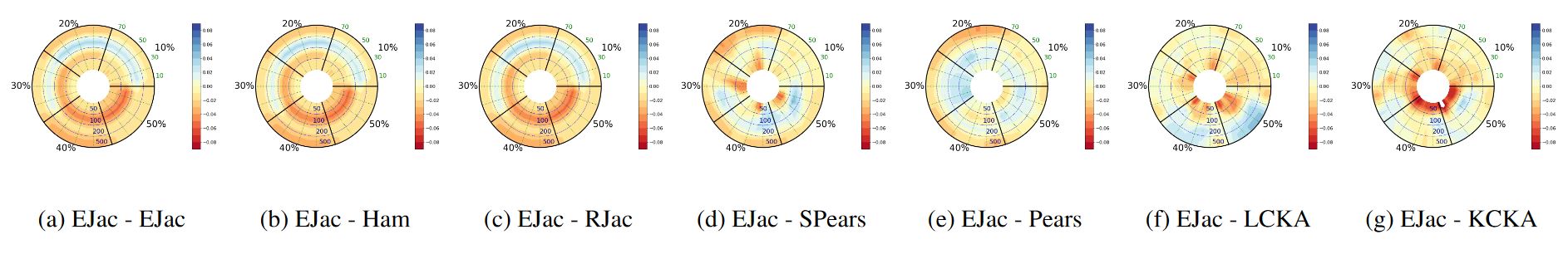}}
    
 \subfloat{
    \includegraphics[width=1 \linewidth]{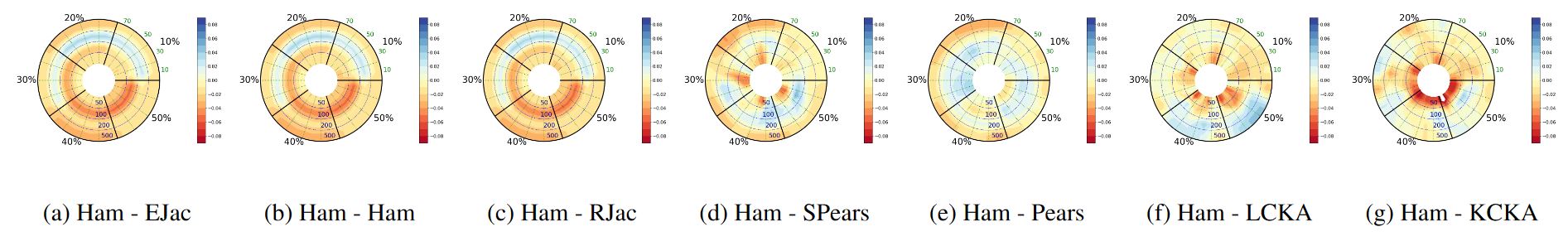}}
    
 \subfloat{
    \includegraphics[width=1 \linewidth]{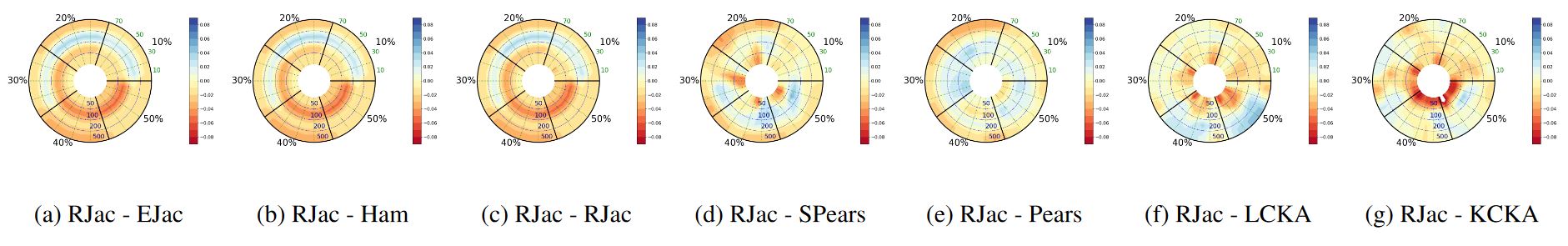}}
    
 \subfloat{
    \includegraphics[width=1 \linewidth]{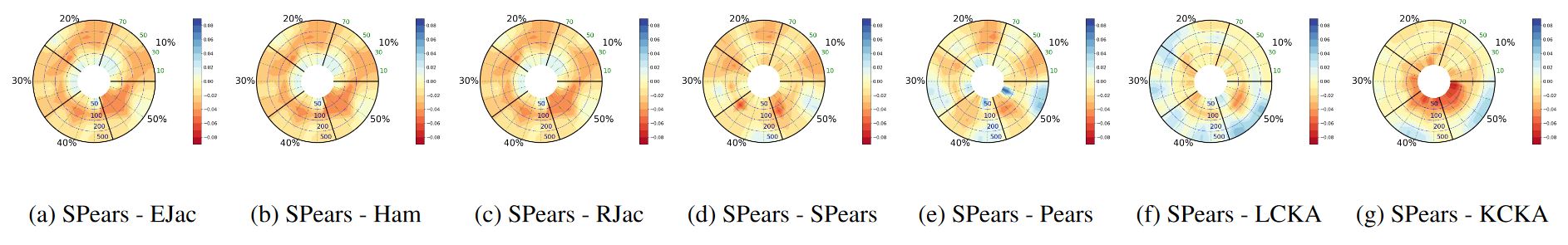}}
    
 \subfloat{
    \includegraphics[width=1 \linewidth]{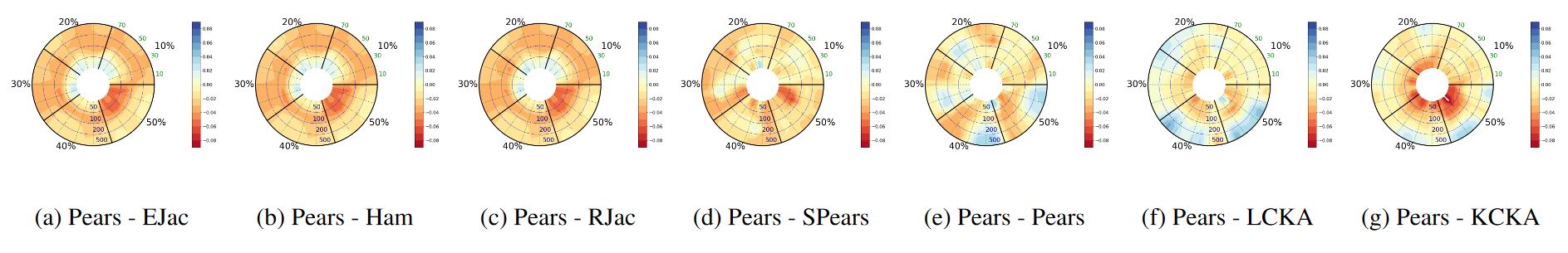}}
    
 \subfloat{
    \includegraphics[width=1 \linewidth]{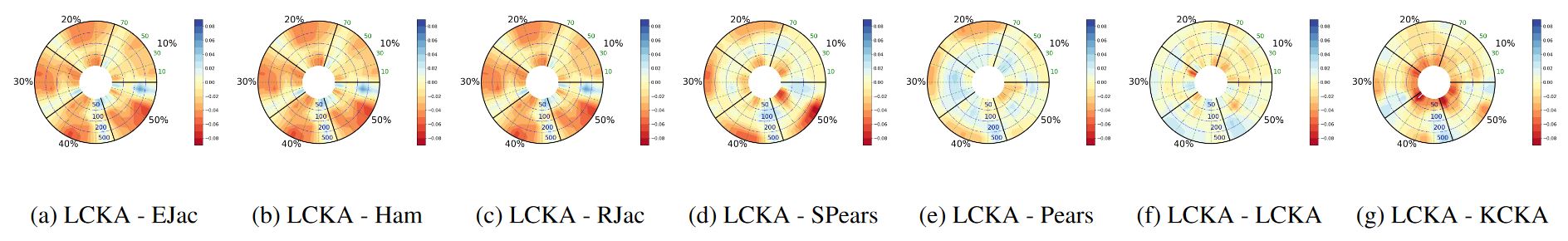}}
    
 \subfloat{
    \includegraphics[width=1 \linewidth]{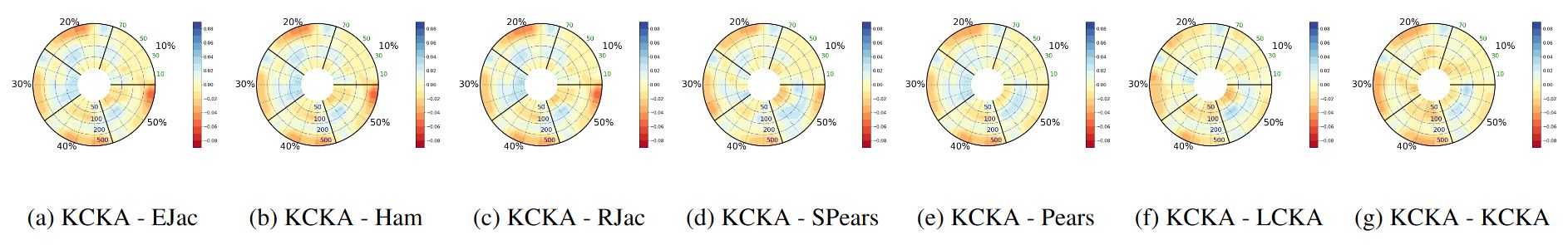}}
    \caption{Hyperparameters' relationship figures for all combinations of distance metrics for MiniPubmed dataset.}
\label{MiniPubmed}
\end{figure*}
\clearpage

\begin{figure*}[h!]
     \centering
 \subfloat{
    \includegraphics[width=1 \linewidth]{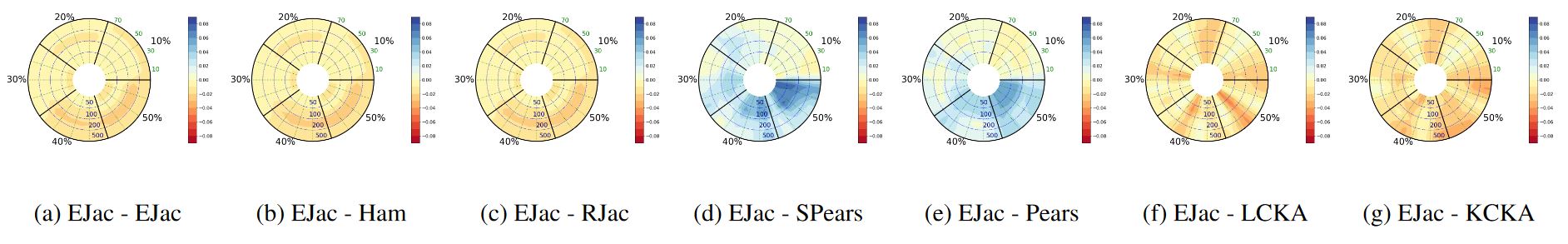}}
    
 \subfloat{
    \includegraphics[width=1 \linewidth]{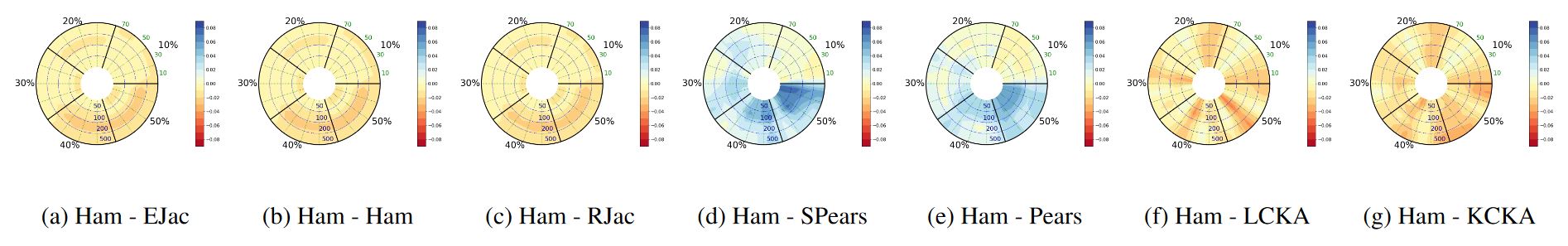}}
    
 \subfloat{
    \includegraphics[width=1 \linewidth]{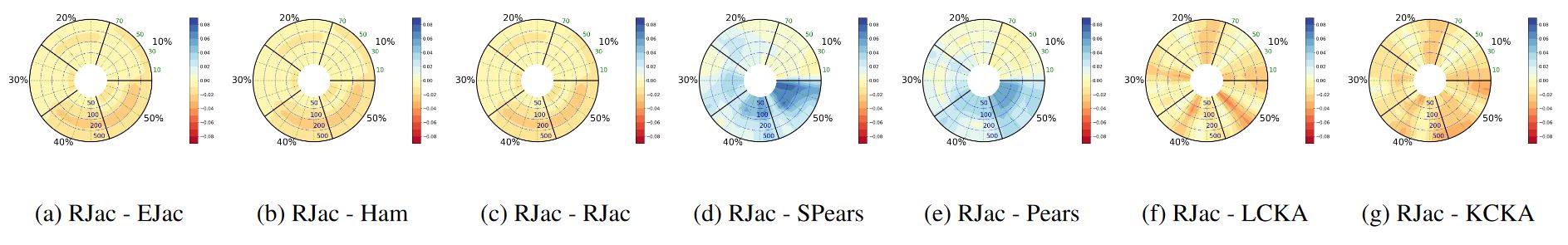}}
    
 \subfloat{
    \includegraphics[width=1 \linewidth]{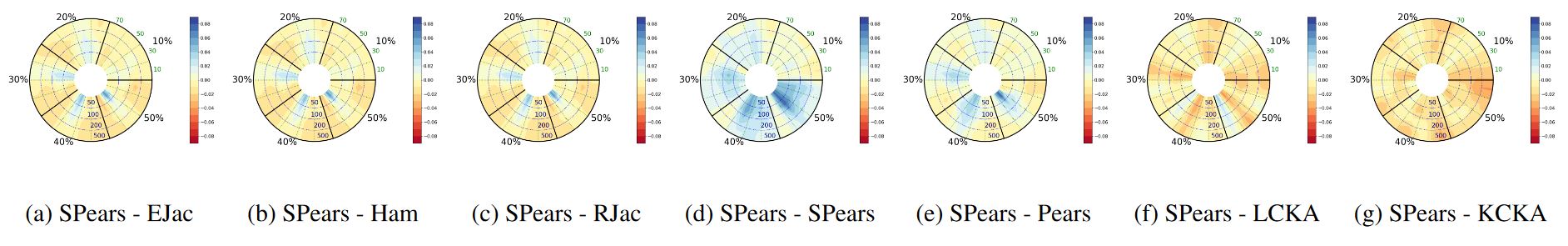}}
    
 \subfloat{
    \includegraphics[width=1 \linewidth]{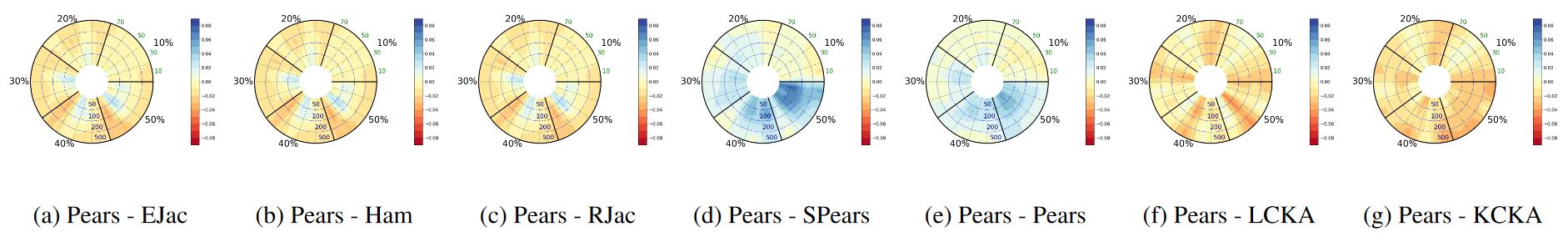}}
    
 \subfloat{
    \includegraphics[width=1 \linewidth]{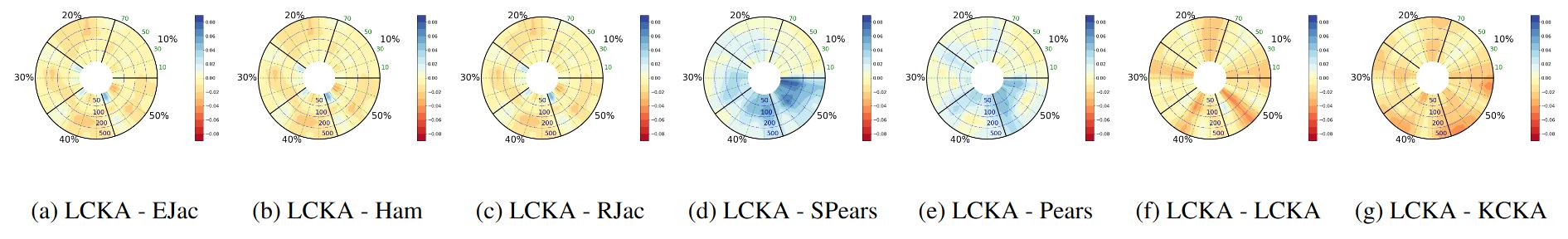}}
    
 \subfloat{
    \includegraphics[width=1 \linewidth]{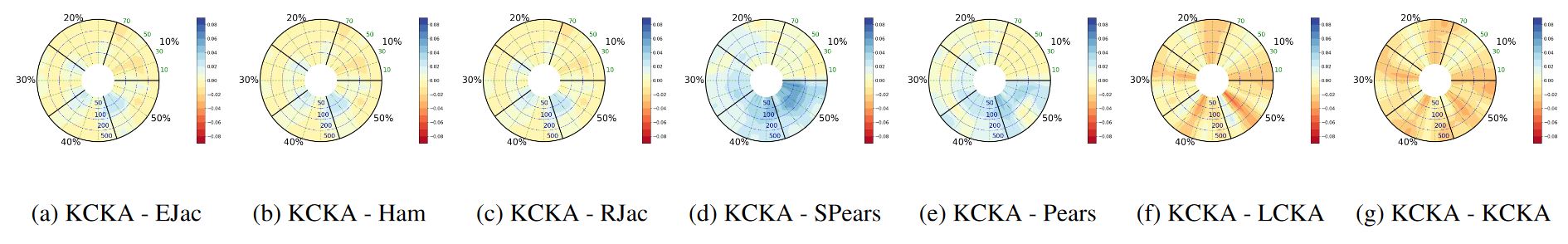}}
    \caption{Hyperparameters' relationship figures for all combinations of distance metrics for Placenta dataset.}
\label{Placenta}
\end{figure*}

\clearpage

\subsection{Semi-Pearson Correlation Coefficients}
Given the representations $H$, and $H'$, the Semi-Pearson Correlation coefficient matrix of these two matrices is as follows.
\begin{align}
    C=\begin{bmatrix}
         R(h_1,h'_1)\;\; 
         R(h_2,h'_2)\;\;
         R(h_3,h'_3)\;\;
    \end{bmatrix}^T
\end{align}
% \begin{align}
%     C=\begin{bmatrix}
%          R(h_1,h'_1)\\
%          R(h_2,h'_2)\\
%          R(h_3,h'_3)\\
%     \end{bmatrix}
% \end{align}
% , where $R(a,b)=\frac{Cov(a,b)}{\sqrt{Var(a).Var(b)}}$ with $Cov(.)$ representing covariance and $Var(.)$ representing variance.
The distance is computed as the average value across all elements of this matrix. The advantage of the Semi-Pearson metric lies in its ability to find the correlation between each node's embedding in one representation matrix and its corresponding one in the other, resulting in a fairer comparison between node representations compared to the Pearson metric, which correlates each node with all others in $H_C$.

%%%%%%%%%%%%%%%%%%%%%%%%%%%%%%%%%%%%%%%%%%%%%%
\section{Additional Hyperparameters' Figures}
\label{HyperParFigs}
This section visually presents the interplay of hyperparameters across various combinations of robustness and performance metrics, as depicted in Figures~\ref{Cora}, \ref{Citeseer}, \ref{MiniPubmed}, and~\ref{Placenta}. 
These figures highlight the distinctions between the original input outcomes and those achieved by the SHERD method, utilizing specific configurations tailored for the Cora, and Citeseer, Mini-Pubmed, and Mini-Placenta datasets, respectively to visualize hyperparameters interplay. %Hyperparameters' interplay visualizations for Mini-Pubmed and Mini-Placenta datasets are available in the Supplementary Material. 
Analogous to Figure~\ref{HypParam-rel}, the blue numbers denoting different radii signify the number of clusters, the green numbers within smaller sectors represent $\tau$ values, while the percentages displayed on larger sectors indicate compression percentages.

%________________________________________________
\section{Conclusion}
This paper introduces SHERD, a novel method for identifying and eliminating vulnerable nodes in node classification using graph neural networks, creating a robust subgraph that preserves performance. By utilizing early training representations to evaluate node susceptibility and informativeness, SHERD efficiently enhances model accuracy and adversarial robustness. Experiments across citation and histology datasets showcase SHERD's versatility and effectiveness in bolstering reliability for crucial graph learning applications through input-level defenses.

As future work, we will delve into the exploration of later representations and their synergies with earlier ones, particularly through the utilization of deeper GCNs. Given that the attacks in this paper encompass various categories of standard attacks, we hypothesis that the outperformance of our method can be extended to many other attacks, which we will explore in future work.

\section*{Acknowledgement}
This work has been supported by NSF CAREER-CCF 2451457; the findings are those of the authors only and do not represent any position of these funding bodies. The authors thank Dariush Bodaghi (former PhD student at UMaine) for the useful discussions.

% \clearpage

\bibliographystyle{IEEEtran}


% \bibliography{egbib}
% \newpage
% \pagebreak
\section*{Biography Section}
\vskip -2\baselineskip plus -1fil
\begin{IEEEbiography}[{\includegraphics[width=1in,height=1.25in,clip,keepaspectratio]{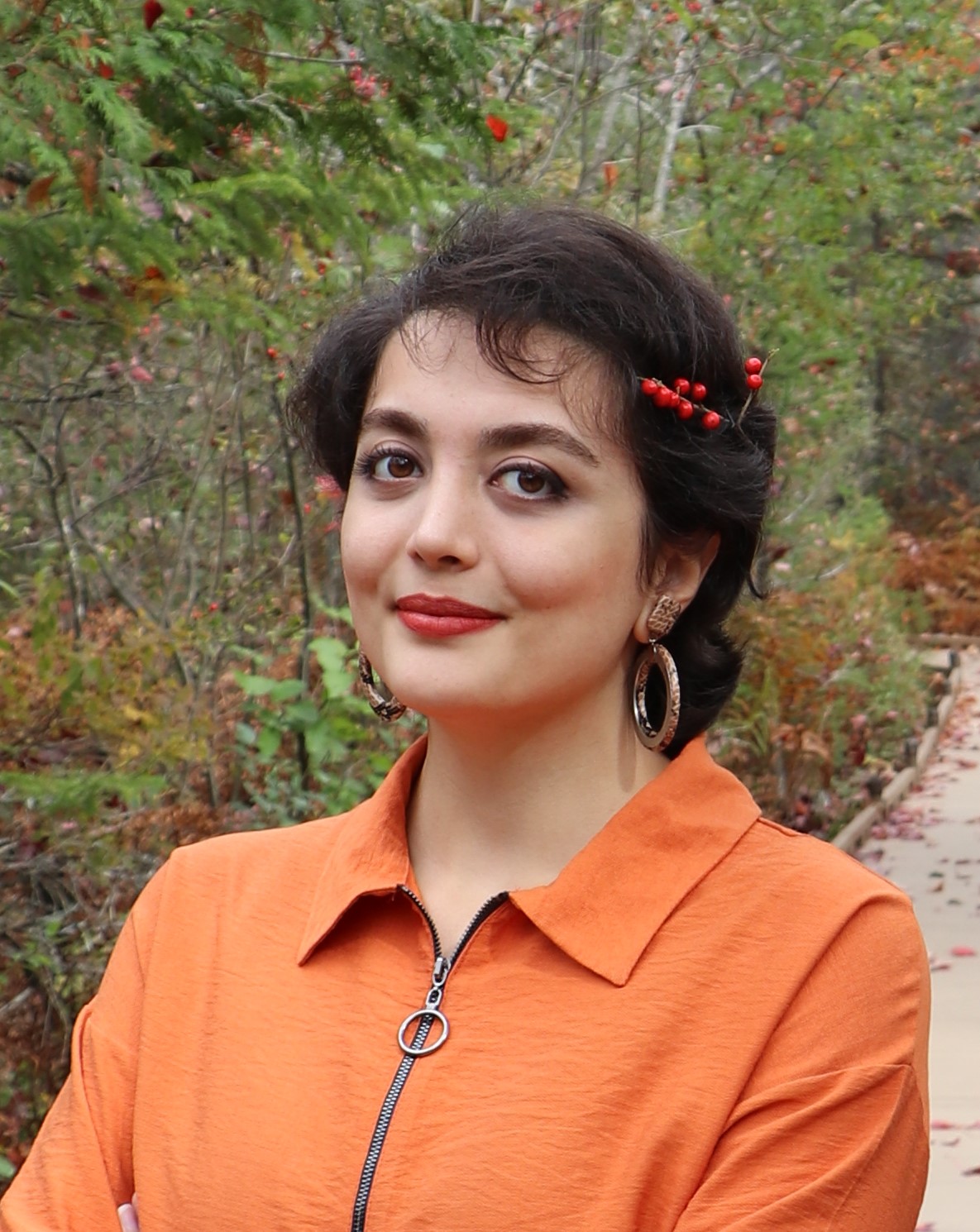}}]{Sepideh Neshatfar} is a Ph.D. student in Computer Science at the University of Maine (UMaine) with the advisory of Prof. Salimeh Yasaei Sekeh since 2021. She was a Teacher Assistant from 2021 to 2022 in the School of Computing and Information Science at UMaine. Since 2023, she has been a Research Assistant in the Sekeh laboratory. She holds a B.S. degree in Computer Engineering (Information Technology) from the Isfahan University of Technology, and a M.S. degree in Computer Science from the University of Maine, U.S.

Mrs. Neshatfar's research interests include graph summarization with a focus on adversarial robustness, computer vision and image processing in medical and autonomous driving applications, and natural language processing.

\end{IEEEbiography}
\vskip -2\baselineskip plus -1fil
\begin{IEEEbiography}[{\includegraphics[width=1in,height=1.25in,clip,keepaspectratio]{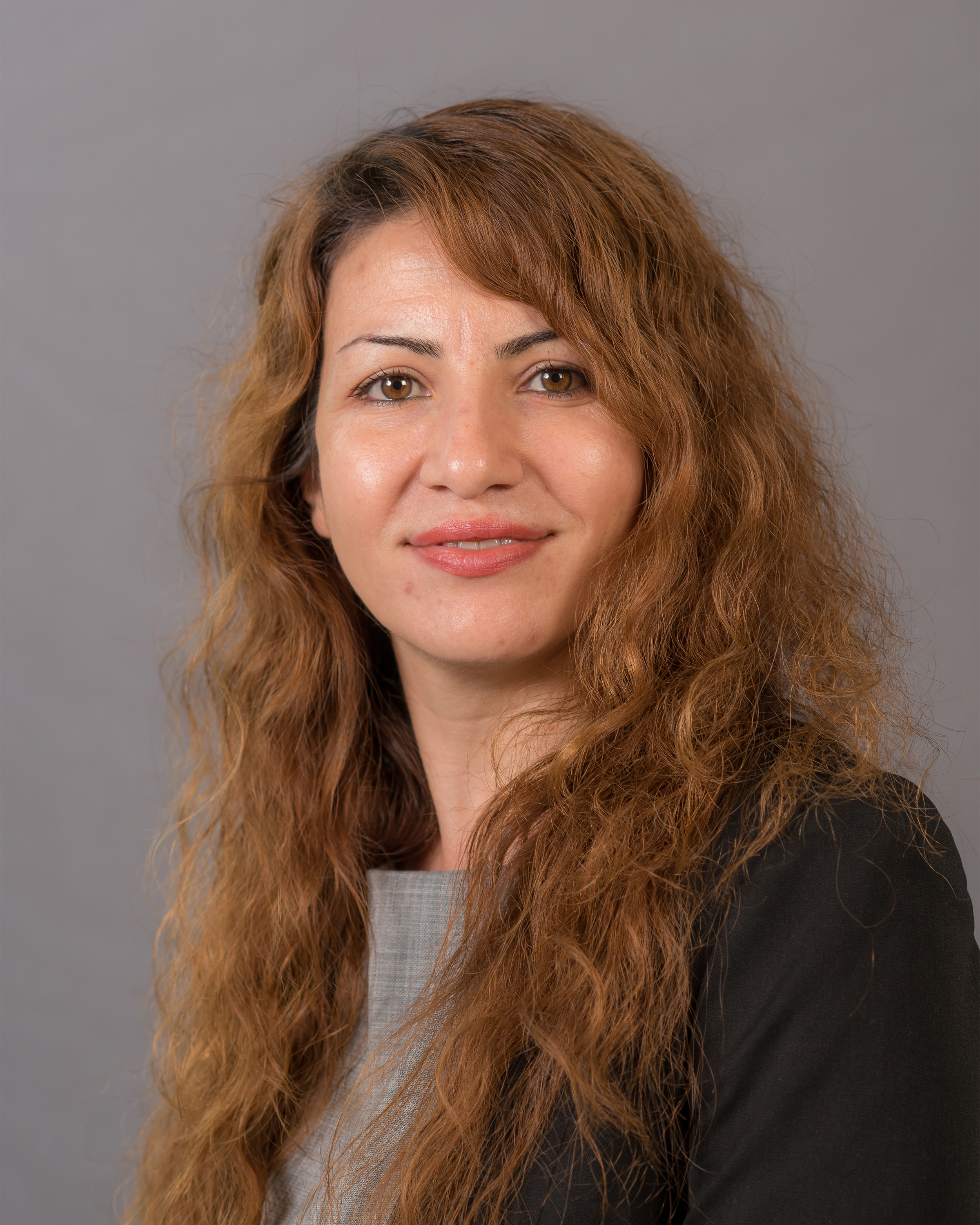}}]{Salimeh Yasaei Sekeh} is an assistant professor in the Computer Science Department of the University of Maine (UMaine). She joined UMaine in 2019 from Michigan, where she was a postdoctoral researcher working with Alfred O. Hero. She held CAPES-PNPD Post-Doctoral Fellow appointment with the Federal University of Sao Carlos (UFSCar), Brazil, in 2014 and 2015. She was a Visiting Scholar with the Polytechnic University of Turin, Turin, Italy, from 2011 to 2013. She was a recipient of CAREER award and CISCO research gift both in 2022. She received the 2023 Maine College of Engineering and Computing (MCEC) Early Career Research Award.

Prof. Sekeh is the Director of the Sekeh Laboratory and her research interests include machine learning, large-scale data science, and computer vision. Her primary focus is in design, improvement, and analysis of deep learning techniques with emphasis on efficiency and robustness. Further, she works on the analysis of continual learning methods, adversarial learning, domain adaptation, graph summarization, and practical applications of machine learning in real-world problems including hyperspectral imaging and forestry-ocean science. 
\end{IEEEbiography}

% \vfill
 % \EOD
\end{document}